\newif\ifarxiv
\arxivtrue

\ifarxiv
  \ifdefined\XeTeXversion\else\pdfoutput=1\fi
  \documentclass{article}
  \usepackage{arxiv}
\else
  \documentclass[twoside]{article}
  \usepackage{aistats2027}
\fi
\usepackage[utf8]{inputenc}
\usepackage[T1]{fontenc}
\usepackage{microtype}
\usepackage{amsmath,amssymb,amsfonts,mathtools}
\usepackage{amsthm}
\usepackage{graphicx}
\usepackage{booktabs}
\usepackage{multirow}
\usepackage{tikz}
\usepackage{pgfplots}
\usetikzlibrary{calc}
\usepgfplotslibrary{groupplots}
\usepackage{float}
\usepackage{nicefrac}
\usepackage[round,authoryear]{natbib}
\ifarxiv
\else

\fi
\usepackage{hyperref}
\usepackage{cleveref}
\usepackage{url}
\usepackage{doi}
\ifarxiv
  \usepackage{axym-links}
  \publicationcode{https://github.com/Axym-Labs/iewc}
  \publicationcorrespondence{dw@axym.org}
\fi

\providecommand{\iewcroot}{}
\definecolor{IEWCPrimary}{HTML}{3F21B6}
\definecolor{IEWCSecondary}{HTML}{8C7AD3}
\definecolor{IEWCDark}{HTML}{3F4147}
\definecolor{IEWCMid}{HTML}{777A80}
\definecolor{IEWCLight}{HTML}{B9BBC0}
\definecolor{IEWCText}{HTML}{20232A}
\definecolor{IEWCGrid}{HTML}{E3E5E8}
\definecolor{IEWCBand}{HTML}{F0F1F3}

\pgfplotsset{
  compat=1.18,
  iewc axis/.style={
    axis x line*=bottom,
    axis y line*=left,
    axis line style={IEWCText,line width=.45pt},
    tick align=outside,
    tick style={IEWCText,line width=.4pt},
    tick label style={font=\scriptsize,text=IEWCText},
    yticklabel style={/pgf/number format/fixed,/pgf/number format/precision=2},
    scale only axis,
    label style={font=\footnotesize,text=IEWCText},
    title style={font=\footnotesize,text=IEWCText,anchor=west,at={(0,1.02)}},
    ymajorgrids,
    grid style={IEWCGrid,line width=.35pt},
    legend style={draw=none,fill=none,font=\footnotesize,/tikz/every even column/.append style={column sep=8pt}},
    legend cell align=left,
    line width=.7pt,
    mark size=1.15pt,
    scaled ticks=false,
    clip=true,
  },
  sequential/.style={color=IEWCLight,mark=*,solid},
  ewc/.style={color=IEWCDark,mark=square*,solid},
  ewcdr/.style={color=IEWCMid,mark=triangle*,dotted},
  iewc/.style={color=IEWCPrimary,mark=*,solid,line width=.85pt},
  iewcrdw/.style={color=IEWCSecondary,mark=diamond*,dashed,line width=.8pt},
  context/.style={color=IEWCLight,mark=none,solid,line width=.6pt},
  target/.style={color=black,mark=none,solid,line width=.85pt},
  accuracy/.style={color=IEWCPrimary,mark=*,solid,line width=.85pt},
  retention/.style={color=IEWCSecondary,mark=square*,dashed,line width=.8pt},
  domainone/.style={color=IEWCPrimary,mark=*,solid,line width=.85pt},
  domaintwo/.style={color=IEWCSecondary,mark=diamond*,dashed,line width=.8pt},
}

\newtheorem{proposition}{Proposition}

\hypersetup{
  hidelinks,
  pdftitle={Improved Elastic Weight Consolidation as an Optimization Constraint for Continual Learning},
  pdfkeywords={continual learning, elastic weight consolidation, empirical Fisher, constrained optimization},
}
\ifarxiv
  \hypersetup{pdfauthor={Davide Wiest}}
\else
  \hypersetup{pdfauthor={}}
\fi

\newcommand{\iewcoverviewcaption}{Each panel shows the continual-learning penalty
  over normalized directions for two representative samples. EWC (OPG) inherits
  sample weighting from output-loss-gradient scale; uniform IEWC removes this
  scale away from the damping regime, while residual-diversity weighting changes
  the aggregation measure as a separate design choice.}

\newcommand{\iewcleadfigure}{%
  \begin{center}
  \refstepcounter{figure}
  \includegraphics[width=0.84\textwidth]{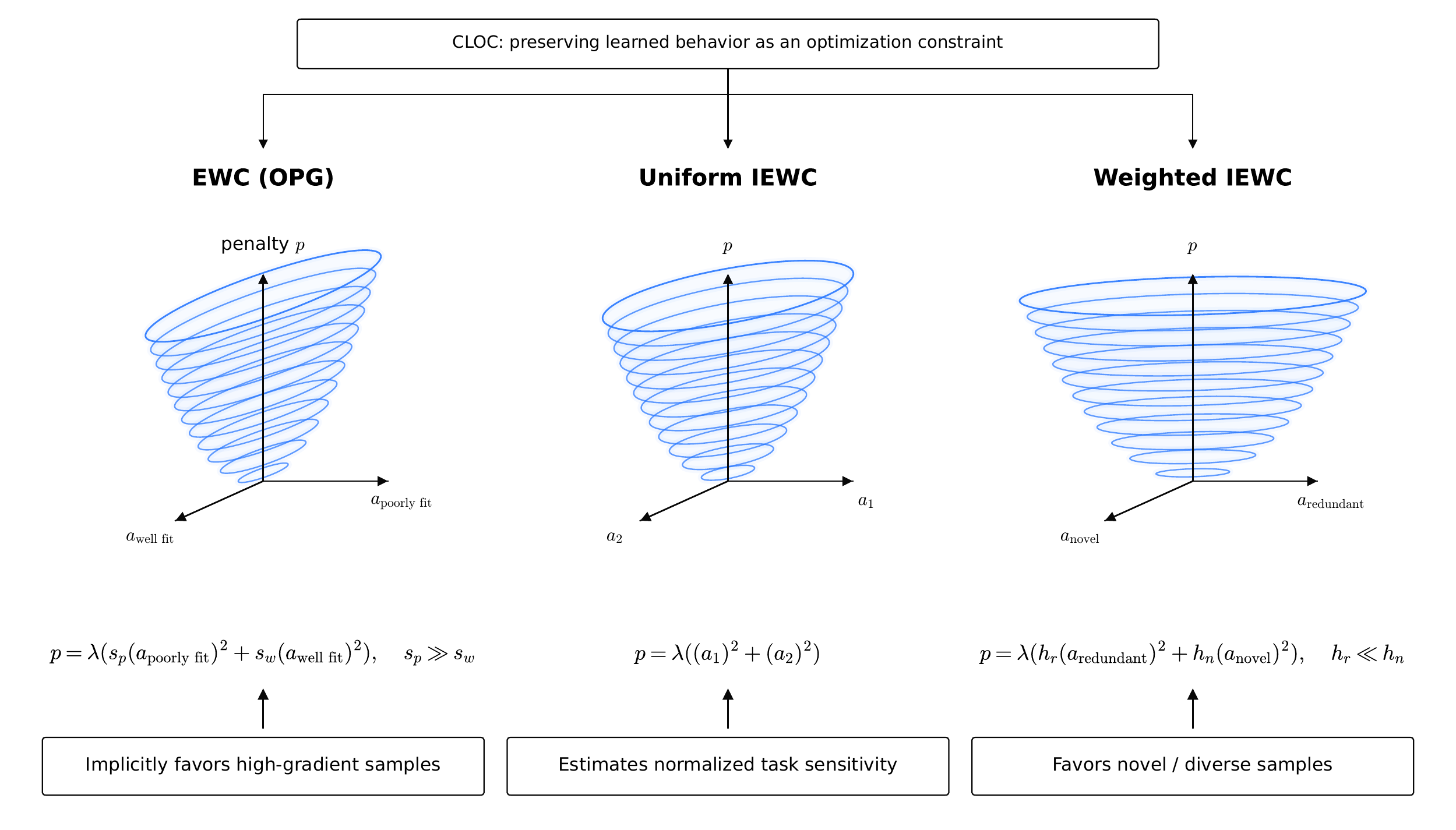}

  \vspace{0.8em}

  \begin{minipage}{0.94\textwidth}
  \small
  \textbf{Figure \thefigure:} \iewcoverviewcaption
  \label{fig:overview}
  \end{minipage}
  \end{center}
  \vspace{0.5em}%
}

\ifarxiv
  \title{Improved Elastic Weight Consolidation as an Optimization Constraint for Continual Learning}
  \author{
    Davide Wiest \\
    Department of Computer Science, TU Darmstadt
  }
  \date{}
  \renewcommand{\headeright}{}
  \renewcommand{\undertitle}{}
  \renewcommand{\shorttitle}{Improved Elastic Weight Consolidation}
\fi

\begin{document}

\ifarxiv
\maketitle
\else
\twocolumn[
\aistatstitle{Improved Elastic Weight Consolidation as an Optimization Constraint for Continual Learning}

\aistatsauthor{Anonymous Author}
\aistatsaddress{Anonymous Institution}

\iewcleadfigure
]
\fi

\begin{abstract}
Regularization-based continual learning reduces catastrophic forgetting by
penalizing changes to parameters that mattered for earlier tasks. Elastic Weight
Consolidation (EWC), a standard instance, weights its quadratic penalty by the
empirical Fisher, usually justified as approximating loss curvature. We
instead derive the penalty from what training on a new task should preserve. In
this formulation, \emph{Continual Learning as an Optimization Constraint}
(CLOC), new training must keep chosen functionals of learned behavior unchanged,
and the local approximation of this constraint is the stored penalty. We show
that EWC's empirical Fisher is exactly the first-order form of this constraint
when the preserved functional is each old sample's loss. The derivation exposes
a flaw: EWC weights each sample by the scale of its loss gradient, so samples
the model fits well barely contribute, although later training can still forget
them. Improved EWC (IEWC) instead penalizes how far each old output moves off
its loss level set, rather than how much its loss changes, which removes this
weighting for any differentiable loss and output geometry. Across
classification, segmentation, language, forecasting, and diffusion, IEWC
variants match or improve on EWC and outperform EWC-DR, a classification-specific
correction. In a held-out evaluation with
validation-selected hyperparameters, removing this weighting alone raises ImageNet-R
accuracy over EWC by 2.00 percentage points (95\% CI 0.93--3.07). More
broadly, CLOC turns the design of continual-learning penalties into a choice of
what to preserve, from which a principled penalty follows by derivation.

\end{abstract}

\ifarxiv
\makepublicationlinks


\begin{figure}[H]
\centering
\includegraphics[width=0.62\textwidth]{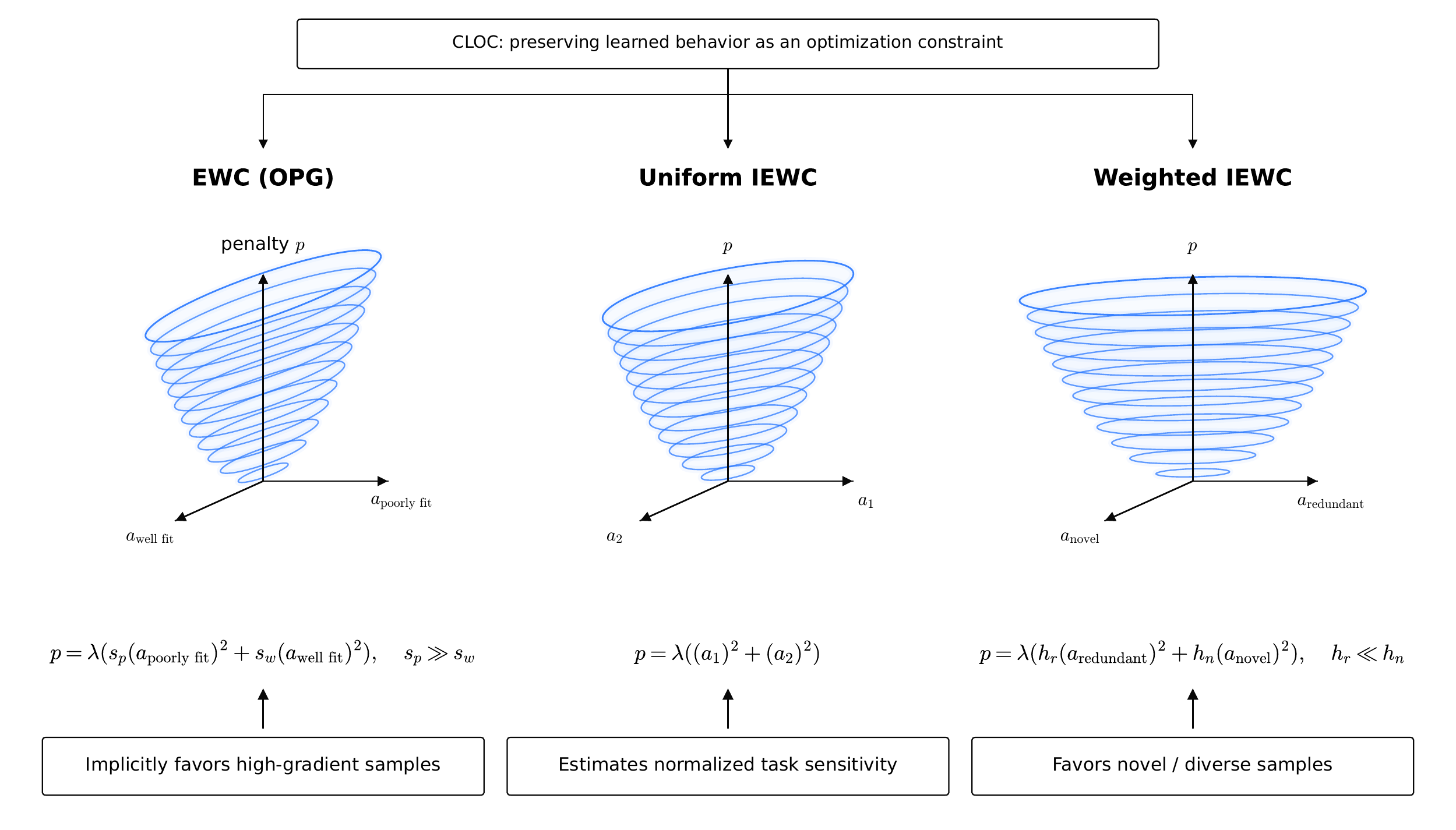}
\caption[IEWC framework overview]{\iewcoverviewcaption}
\label{fig:overview}
\end{figure}
\fi

\section{Continual Learning as an Optimization Constraint}

Continual learning asks a model to learn from sequentially available distributions without unrestricted access to earlier data \citep{McCloskeyCohen1989,French1999,Parisi2019,VanDeVenTolias2019}. After training on distribution $A$ and obtaining $\theta_A^\star$, minimizing only the later loss $L_B(\theta)$ leaves previously learned behavior unconstrained. Methods such as EWC, SI, MAS, and LwF address this problem with different preservation objects: parameter importance, learning-path importance, output sensitivity, or old predictions \citep{EWC,SI,MAS,LwF}. This variety raises a prior question: \emph{what quantity should a local continual-learning penalty preserve?}

We pose that question directly as constrained optimization. Let $q:\Theta\rightarrow\mathcal H$ be a quantity whose old value or acceptable set represents preservation, and let $\mathcal C\subset\mathcal H$ encode the corresponding constraint. Learning distribution $B$ becomes
\[
\min_{\theta\in\Theta} L_B(\theta)
\qquad
\text{subject to}
\qquad
q(\theta)\in\mathcal C.
\]
We call this perspective \emph{Continual Learning as an Optimization Constraint} (CLOC). CLOC is a formulation rather than a single algorithm: different choices of $q$, $\mathcal C$, relaxation, and approximation recover different local preservation penalties.

The paper develops one path through this design space. First, a population and low-order relaxation turns samplewise preservation objects into a stored quadratic penalty. Second, preserving any differentiable scalar functional yields a gradient outer-product matrix; choosing the old-sample loss gives OPG, and negative log likelihood gives the empirical Fisher used in common EWC implementations. This derivation shows that EWC weights each old sample by the squared scale of its output-loss gradient, so samples the model already fits well contribute little to the penalty. Third, preserving output motion only in the normal quotient of the old loss level set yields the iEF normalization and IEWC, which removes this weighting. We make damping, sample weighting, output metric, and matrix approximation explicit design choices instead of treating them as consequences of the normalization. Because the derivation comes first, it makes predictions that the experiments test: where output-loss-gradient scales vary across samples, OPG importance spans a much wider range than IEF importance, and high-loss corrupted samples dominate OPG but not IEF (Section~\ref{sec:empirical}, Appendix~\ref{app:contamination}).

Concretely, we contribute (i) a functional-preservation construction whose loss specialization derives empirical-Fisher EWC and level-set-normalized IEWC within one framework; (ii) a matched output-metric construction and the exact finite-damping, interpolation, and loss-unit boundary; and (iii) a direct spectral test of the mechanism, results across five domains, and a matched held-out evaluation with validation-selected hyperparameters in which IEWC improves ImageNet-R accuracy over EWC by 2.00 percentage points.

\begin{table}[t]
\caption{Core notation; Appendix~\ref{app:notation} gives the complete table.}
\label{tab:core-notation}
\centering
\small
\begin{tabular}{@{}lp{0.68\columnwidth}@{}}
\toprule
Symbol & Meaning \\
\midrule
$J_i,r_i,g_i$ & Output Jacobian, output-loss gradient, and $g_i=J_i^\top r_i$. \\
$M_{\mathrm{OPG}}$ & Mean $g_i g_i^\top$ from scalar preservation. \\
$s_i=r_i^\top G_i^{-1}r_i$ & Squared dual norm of the output-loss gradient. \\
$M_{\mathrm{IEF}}$ & Mean $g_i g_i^\top/(s_i+\tau_A)$ from level-set preservation. \\
\bottomrule
\end{tabular}
\end{table}

\section{Related Work and Scope}
\label{sec:related-work}

EWC stores a quadratic approximation around earlier parameters and weights parameter motion by a Fisher-derived importance estimate \citep{EWC}. Other parameter-importance regularizers estimate importance from the optimization trajectory, output sensitivity, or a combination of EWC-style curvature and path-integral evidence \citep{SI,MAS,RWalk}; their apparently different estimators have also been analyzed within a common framework \citep{BenzingRegularisation}. Bayesian and posterior-combination approaches instead accumulate task posteriors or merge their moments \citep{ProgressCompress,VCL,IMM}. A practical ambiguity remains: implementations called ``Fisher'' may use the model-label expectation, sampled model labels, observed labels, or minibatch aggregates. These estimators are not interchangeable in theory or practice \citep{KunstnerEF,VanDeVenFisher2025}. Our scalar-loss derivation concerns the observed-label OPG object; sampled model-Fisher EWC is evaluated separately rather than treated as another name for the same matrix.

Regularization can instead protect functions or representations. LwF distills old predictions on later inputs, encoder-based lifelong learning constrains reconstructions of old-task features, and functional regularization preserves selected predictive functions or memorable points \citep{LwF,EBLL,FunctionalGP,FROMP}. More generally, many training objectives contain vectors or matrices that are computed from parameters or activations but held constant with respect to the parameters being optimized, such as weight-averaged teacher parameters \citep{MeanTeacher}, stop-gradient target projections \citep{BYOL,SimSiam}, predictions of an earlier model \citep{LwF}, and running activation moments \citep{TeReL}. EWC and IEWC follow the same pattern in the sense that both compute an importance matrix once earlier tasks are learned and then hold it fixed as the weighting of a quadratic penalty. CLOC complements this taxonomy: it begins with a preservation specification and asks which local stored penalty follows. The algebra below allows differentiable scalar functionals, while IEWC specializes to scalar-loss preservation and loss-level-set-normal output preservation.

iEF was introduced as a normalized observed-gradient approximation for natural-gradient optimization \citep{IEF}. We recover its Euclidean algebra from a different question: the metric distance of an old output displacement from the tangent to its old loss level set. This gives iEF a preservation interpretation that does not rely on curvature. Our additions relative to simply substituting iEF into EWC are the CLOC derivation, the matched output-metric generalization, the finite-damping and interpolation boundary, and the separation of normalization from aggregation weights. EWC-DR observes in classification that well-fitted samples lose importance under EWC and repairs this with logit reversal \citep{EWC-DR}; CLOC identifies output-loss-gradient scale as the general cause, and IEWC removes it for any differentiable loss and output geometry. The RDW ablation borrows a streaming residual-diversity idea from GSS \citep{GSS}; it is an order-dependent aggregation heuristic, separate from the level-set result.

\section{Population Setting and Approximations}

For a closed convex set $\mathcal C$, the Moreau envelope of its indicator is a scaled squared distance \citep{MoreauProx}. This motivates the soft CLOC objective
\[
\begin{aligned}
\min_\theta J(\theta)
&=\min_\theta\left[
L_B(\theta)
+\frac{\lambda}{2}\operatorname{dist}(q(\theta),\mathcal C)^2
\right]\\
&=\min_\theta\left[
L_B(\theta)
+\frac{\lambda}{2}\inf_{u\in\mathcal C}\|q(\theta)-u\|^2
\right].
\end{aligned}
\]
GEM and A-GEM are similar in spirit in that they formulate preservation through constraints on old examples \citep{GEM,AGEM}. CLOC turns the old-task constraint into a penalty built after task-$A$ training. In a population setting, we use sample-based analogues $\ell_B$ and $q$ and take the empirical average. Define the unscaled preservation loss as $p(a,\theta)=\operatorname{dist}(q(a,\theta),\mathcal C)^2$.
\[
\min_\theta
\;
\frac{1}{N_B}\sum_{i=1}^{N_B}\ell_B(b_i,\theta)
+
\frac{\lambda}{2N_A}\sum_{i=1}^{N_A}p(a_i, \theta)
\]
Additional old distributions enter through analogous constraints. We refer to this population version as \textbf{population CLOC (P-CLOC)}. Efficiency depends on the penalty $p$. Ideally, the constraint contributes its gradient to each gradient step. However, recomputing the full old-distribution penalty $p$ and its associated factors many times during optimization on distribution $B$ is usually prohibitively expensive, since it depends on the dataset $D_A$ associated with $A$. Reusing $D_A$ later also violates the continual-learning setting. For this reason, we replace the exact penalty by a local $k$-th order Taylor-polynomial approximation around the old solution $\theta_A^\star$.
This turns the computation of $p$ into a repeated evaluation of precomputed derivative tensors and bounds memory requirements to $O(d^k)$ for a $d$-dimensional parameterization. Since memory scales exponentially with $k$, the standard scalable surrogate is a low-order Taylor approximation. We refer to this local approximation as \textbf{low-order CLOC (L-CLOC)}, and to its population version as \textbf{PL-CLOC}. It remains to specify the preserved quantity $q$ and constraint set $\mathcal{C}$. Because each term of $p$ depends on one sample, $p$ and its gradients can be analyzed per sample.

\subsection{Scalar Functional Preservation}
\label{sec:scalar-functional-preservation}

Let $s_a:\Theta\to\mathbb R$ be a differentiable scalar functional of the model on old sample $a$. Preserving that functional means
\[
q(a,\theta)
=
s_a(\theta)-s_a(\theta_A^\star).
\]
Since preservation means $q(a,\theta)=0$, take $\mathcal C=\{0\}$. The corresponding unscaled proximal loss is
\[
p(a,\theta)
=
\operatorname{dist}(q(a,\theta),\mathcal C)^2
=
q(a,\theta)^2.
\]
To obtain a low-order Taylor approximation, use the perturbation around the old solution $\delta=\theta-\theta_A^\star$ in the first-order expansion of $q$:
\[
q(a,\theta_A^\star+\delta)
\approx
g_a^\top\delta,
\]
where
\[
g_a
=
\nabla_\theta s_a(\theta)\big|_{\theta=\theta_A^\star}.
\]

When the functional factors through the model output as $s_a(\theta)=\phi_a(f_\theta(x))$, write
\[
\begin{aligned}
z_{A,i} &= f_{\theta_A^\star}(x_i),\\
r_i &= \nabla_z \phi_{a_i}(z)\big|_{z=z_{A,i}},\\
J_i &= \nabla_\theta f_{\theta_A^\star}(x_i),
\end{aligned}
\]
the chain rule gives
\[
g_i
=
J_i^\top r_i.
\]

Substituting this expansion into the preservation penalty gives
\[
\frac{\lambda}{2N_A}\sum_{i=1}^{N_A} q(a_i,\theta)^2
\;\approx\;
\frac{\lambda}{2N_A}\sum_{i=1}^{N_A} (g_i^\top\delta)^2.
\]
Collecting the gradient outer products yields
\[
\frac{\lambda}{2N_A}\sum_{i=1}^{N_A} (g_i^\top\delta)^2
=
\frac{\lambda}{2}\,
\delta^\top
\left(
\frac{1}{N_A}\sum_{i=1}^{N_A} g_i g_i^\top
\right)
\delta
=
\frac{\lambda}{2}\,
\delta^\top
M_1
\delta.
\]

Therefore the resulting matrix is
\[
M_1
=
\frac{1}{N_A}\sum_{i=1}^{N_A} g_i g_i^\top
=
\frac{1}{N_A}\sum_{i=1}^{N_A} J_i^\top r_i r_i^\top J_i.
\]
The construction is exact for the squared first-order surrogate displayed above, not for the original nonlinear equality constraint. The rest of the paper takes $\phi_a=\ell_A$ unless stated otherwise. This loss specialization is the \emph{outer-product loss-gradient} (OPG) matrix, and for negative log likelihood it becomes the empirical-Fisher algebra used by EWC, as specified next.

\subsection{Outer-Product Loss-Gradient Regularization and EWC}

For the loss functional $s_{a_i}(\theta)=\ell_A(f_\theta(x_i))$, the matrix obtained in the previous section is
\[
M_{\mathrm{OPG}}
=
\frac{1}{N_A}\sum_{i=1}^{N_A}g_i g_i^\top.
\]
It is an outer product of loss gradients. This identity holds for any differentiable loss, without a probabilistic model or a curvature interpretation. We therefore call it an \emph{outer-product-of-loss-gradients} (OPG) matrix.

For a negative-log-likelihood loss,
\[
\ell_i(\theta)=-\log p_\theta(y_i\mid x_i),
\]
we have $g_i=-\nabla_\theta\log p_\theta(y_i\mid x_i)$. The same OPG algebra then coincides with the observed-label score-gradient matrix commonly called the empirical Fisher,
\[
F_{\mathrm{emp}}
=
\frac{1}{N_A}\sum_{i=1}^{N_A}
\nabla_\theta\log p_\theta(y_i\mid x_i)
\nabla_\theta\log p_\theta(y_i\mid x_i)^\top.
\]
Thus OPG-EWC and empirical-Fisher EWC denote the same implementation in an explicit likelihood setting. The CLOC derivation identifies this matrix as the quadratic form of a squared first-order loss-preservation surrogate, an interpretation distinct from model-Fisher and Hessian curvature.

Writing $s_\theta(x,y)=\nabla_\theta\log p_\theta(y\mid x)$, the model Fisher is instead
\[
F_{\mathrm{model}}(\theta)
=\mathbb E_{x\sim\mathcal D_A}
\mathbb E_{y\sim p_\theta(\cdot\mid x)}
\bigl[s_\theta(x,y)s_\theta(x,y)^\top\bigr].
\]
It averages over labels sampled from the model, not only the observed label. Under standard regularity conditions this Fisher is connected to expected negative-log-likelihood curvature and natural-gradient geometry \citep{Amari1998,MartensNG,KFAC}. The observed-label empirical Fisher is a different object and can be a poor curvature approximation \citep{KunstnerEF}; Fisher-estimator choice can also materially affect EWC behavior and regularization scale \citep{VanDeVenFisher2025}.

Our experiments include a one-label Monte Carlo model-Fisher baseline, obtained by sampling from the frozen predictive distribution at each supervised output site (one class per image and one token per supervised TRACE answer position), alongside OPG-EWC.

\section{Level-Set Preservation in Output Space}
\label{sec:level-set-preservation}

Section~\ref{sec:scalar-functional-preservation} preserves a scalar functional value directly. A second construction instead preserves the model output up to motions along a level set of a differentiable output functional $\phi_a:\mathcal Z\to\mathbb R$. This output-space perspective connects to distillation and functional regularization approaches to continual learning \citep{LwF,Distillation,FunctionalGP,FROMP}. Let $a \in D_A$ be an old sample, write $z_A=f_{\theta_A^\star}(x)$, and define the exact preservation set
\[
\mathcal C_a=
\{z:\phi_a(z)=\phi_a(z_A)\}.
\]
The preserved quantity and hard constraint are, respectively,
\[
q(a,\theta)=f_\theta(x),
\qquad
q(a,\theta)\in\mathcal C_a,
\]
and the corresponding unscaled proximal loss is
\[
p(a,\theta) =
\operatorname{dist}_{G}\big(q(a,\theta),\mathcal C_a\big)^2,
\]
where $G\succ0$ is the chosen local output metric. Positive definiteness makes the primal metric and its inverse on covectors unambiguous; Section~\ref{sec:output-geometry} treats the restricted pseudoinverse convention needed by a singular ordered-output example.

This exact penalty is generally not tractable. To obtain a practical form, use the standard first-order local model of the constraint set near the old output $z_A$, namely its tangent plane. Let
\[
r=
\nabla_z \phi_a(z)\big|_{z=z_A}.
\]
We first assume $r\ne0$, so the functional's level set has a regular tangent hyperplane at $z_A$.
For a small output displacement $\Delta z=f_\theta(x)-f_{\theta_A^\star}(x)$, its first-order change is approximately $r^\top \Delta z$.
Hence the local constraint set is
\[
T=
\{\Delta z:r^\top \Delta z=0\}.  
\]
Replacing the exact level set by this tangent hyperplane gives the local proximal loss
\[
p(a,\theta)  
\approx  
\operatorname{dist}_{G}(\Delta z,T)^2.  
\]

For the metric $|v|_G^2=v^\top G v$, the squared distance to the tangent hyperplane follows from projecting $\Delta z$ onto the gradient:
\[
\operatorname{dist}_{G}(\Delta z,T)^2 =
\frac{(r^\top \Delta z)^2}{r^\top G^{-1}r}.  
\]
Thus $G$ measures primal output displacement while $G^{-1}$ acts on the loss covector; Appendix~\ref{app:matched-metric-projection} gives the projection calculation establishing this matched pair.
Linearizing the model output around $\theta_A^\star$, with
\[
\delta=\theta-\theta_A^\star,  
\qquad  
J=\nabla_\theta f_{\theta_A^\star}(x),  
\]
we have
\[
\Delta z \approx J\delta.  
\]
Substituting this into the local regularization term gives
\[
\frac{\lambda}{2}p(a,\theta_A^\star+\delta)  
\approx  
\frac{\lambda}{2}  
\frac{(r^\top J\delta)^2}{r^\top G^{-1}r}.  
\]
Equivalently,
\[
\frac{\lambda}{2}p(a,\theta_A^\star+\delta)  
\approx  
\frac{\lambda}{2}  
\delta^\top  
\left(
J^\top  
\frac{rr^\top}{r^\top G^{-1}r}  
J  
\right)
\delta.  
\]

Taking the empirical average over old-distribution samples $a_1,\dots,a_N$ yields an importance matrix of the form
\[
M_{2,A} =
\frac{1}{N}\sum_{i=1}^N  
J_i^\top  
\frac{r_i r_i^\top}{r_i^\top G_i^{-1}r_i}  
J_i.  
\]

This is a local construction: it penalizes motion normal to the tangent model, not distance to a globally convex neural-network level set. IEWC is the specialization $\phi_a=\ell_A$ studied below; other functionals define further CLOC instances. When $r_i=0$, the tangent normal is undefined and the undamped quotient is $0/0$; our finite-damping implementation assigns that sample zero contribution. Section~\ref{sec:well-fitted} makes the resulting near-interpolation boundary explicit.

\subsection{Level-Set Normalization and the Improved Empirical Fisher}

For the loss-functional specialization $\phi_a=\ell_A$ and Euclidean output metric $G_i=I$, the matrix from the previous section is the improved empirical Fisher (iEF) introduced by Wu et al. \citep{IEF}:
\[
M_{2,I}=F_{\mathrm{IEF}}.
\]
Hence iEF is the quadratic form of a first-order local constraint that protects old behavior geometrically: instead of preserving a raw scalar-loss coordinate, it penalizes only output motion normal to the old loss level set. The original iEF paper derives the same normalization for approximate natural-gradient descent. In that setting, the observed-label OPG reweights sample contributions by squared output-gradient magnitude; iEF removes this scale factor and retains the normalized output-loss direction. Here the normalization arises from metric projection onto the normal quotient of the old-loss tangent space.

For scalar least-squares regression in Euclidean geometry, Appendix~\ref{app:scalar-gauss-newton} shows that undamped IEF is exactly the Gauss--Newton or output-Jacobian matrix \citep{GaussNewton,MartensHF,MartensNG,IEF}. This matches our geometric intuition: loss preservation is governed by how parameters move the output, which is encoded by the Jacobian $J$, not by the current output-gradient magnitude.

\subsection{Normalization and Sample Weighting}
\label{sec:normalization-weighting}

To compare both matrices in a common notation, let $G_n\succ0$ denote the chosen output metric for sample $n$. Although $M_{\mathrm{OPG}}$ itself does not depend on $G_n$, its summands can be rewritten relative to that metric. In a negative-log-likelihood setting, $M_{\mathrm{OPG}}=F_{\mathrm{emp}}$.

Define
\[
\hat r_n^{\,G}
=
\frac{r_n}{\sqrt{r_n^\top G_n^{-1}r_n}},
\quad
a_n^{\,G}
=
J_n^\top \hat r_n^{\,G},
\quad
w_n^{\,G}
=
r_n^\top G_n^{-1}r_n.
\]
Then, wherever $r_n\ne0$,
\[
\begin{aligned}
M_{\mathrm{IEF},G}
&=
\frac{1}{N}\sum_{n=1}^N a_n^{\,G}(a_n^{\,G})^\top,
\\
M_{\mathrm{OPG}}
&=
\frac{1}{N}\sum_{n=1}^N w_n^{\,G}\,a_n^{\,G}(a_n^{\,G})^\top.
\end{aligned}
\]

Both matrices empirically aggregate the same local old-task sensitivity directions $a_n^{\,G}(a_n^{\,G})^\top$, but they differ in how strongly each sample contributes. Undamped IEF keeps uniform empirical weighting of these normalized directions, while OPG multiplies each contribution by the metric-dependent squared magnitude
\[
w_n^{\,G}=r_n^\top G_n^{-1}r_n.
\]

Thus normalization changes sample aggregation: OPG shifts old-task weighting toward samples with large output-loss gradients and away from confidently fitted ones, whereas undamped IEF removes that scale. This gives the prediction tested in Section~\ref{sec:empirical}: where $w_n^{\,G}$ varies widely across samples, OPG summands spread over a much wider range than IEF summands.

The next section treats probabilistic classification, where the scale difference has a concrete well-fitted limit.

\section{Gradient-Scale Behavior Near Well-Fitted Solutions}
\label{sec:well-fitted}

The output-gradient weighting in observed-label OPG has an important classification symptom. For cross-entropy with softmax probabilities $p_n$ and one-hot label $y_n$, the output-loss gradient with respect to logits is
\[
r_n=p_n-y_n.
\]
For a confidently correct prediction with class $c$ and small $\varepsilon$,
\[
p_{n,c}=1-\varepsilon,
\qquad
\|r_n\|_2^2\in\Theta(\varepsilon^2).
\]
Thus the OPG contribution becomes small as the model becomes more confident and correct. This is coherent with its derivation: the scalar first-order loss change is also small. For continual-learning preservation, however, small current loss need not mean that losing the sample later is harmless. EWC-DR identifies this classification symptom and proposes a logit-reversal repair \citep{EWC-DR}.

This observation is one instance of the weighting effect identified in Section~\ref{sec:normalization-weighting}: observed-label OPG importance depends on output-gradient scale. Undamped IEF removes this particular factor. Positive damping retains the removal above its transition scale, and the controlled comparisons below measure the empirical effect.

The same observation extends to arbitrary differentiable losses.

Importance is estimated after training on $A$, when many old samples may be well fit. The original EWC Laplace argument likewise expands at an old-task optimum, although its model-Fisher motivation differs from the observed-label OPG analyzed here \citep{EWC,KunstnerEF}. Modern expressive models can fit training data up to interpolation \citep{ModernInterp}, so the near-zero regime must be explicit. For one sample, let $s_n=r_n^\top G_n^{-1}r_n$ and
\[
S_n^{\mathrm{OPG}}=J_n^\top r_nr_n^\top J_n,
\qquad
S_{n,\tau}^{\mathrm{IEF},G}=\frac{J_n^\top r_nr_n^\top J_n}{s_n+\tau}.
\]

\begin{proposition}[Gradient-scale behavior near a well-fitted solution]
\label{prop:well-fitted-scaling}
Assume $m_GI\preceq G_n\preceq M_GI$ and
$\|J_n^\top v\|_2\le C_J\|v\|_2$. For $s_n>0$,
\[
S_n^{\mathrm{OPG}}
=s_nS_{n,0}^{\mathrm{IEF},G},
\qquad
\begin{aligned}
\|S_n^{\mathrm{OPG}}\|_2&\in O(s_n),\\
\|S_{n,0}^{\mathrm{IEF},G}\|_2&\in O(1).
\end{aligned}
\]
If $\|J_n^\top r_n/\sqrt{s_n}\|_2\ge c_J>0$, these rates are
$\Theta(s_n)$ and $\Theta(1)$. Consequently, for Euclidean softmax
cross-entropy with correct-class probability $1-\varepsilon$, they are
$\Theta(\varepsilon^2)$ and $\Theta(1)$, respectively. For $\tau>0$,
\[
S_{n,\tau}^{\mathrm{IEF},G}
=
\frac{s_n}{s_n+\tau}S_{n,0}^{\mathrm{IEF},G},
\]
so fixed positive damping restores attenuation and
$S_{n,\tau}^{\mathrm{IEF},G}\to0$ as $r_n\to0$. Finally,
\[
\tau_A
=
\kappa\operatorname{median}\{s_i:s_i>0\}
\]
makes every damped summand invariant to a global positive rescaling of the task loss.
\end{proposition}
Appendix~\ref{app:well-fitted-proof} gives the proof.

The proposition isolates the central EWC--IEWC difference. OPG weights each normalized parameter-sensitivity direction by the squared dual norm $s_n$ of the current output-loss gradient; undamped IEF removes that factor, although model sensitivity through $J_n$ remains. Thus a small current loss gradient need not make the corresponding old behavior unimportant for continual learning.

Finite damping introduces a transition rather than uniformly removing gradient scale: normalization applies when $s_n\gg\tau$, while $s_n\ll\tau$ again suppresses the sample. At exact interpolation we define the contribution as zero; the undamped expression is $0/0$ and has no canonical value without a directional limit. We use $\kappa=10^{-2}$ in the held-out evaluation. Relative damping is invariant only to global loss-unit changes, not arbitrary sample-dependent transformations; if all recorded $s_i$ vanish, the task importance is the zero matrix.

\section{Output-Space Geometry}
\label{sec:output-geometry}

The scalar CLOC construction and its OPG matrix require no output-space metric. The level-set construction does: for a symmetric positive-definite primal metric $G_n$, output displacements have inner product $\langle u,v\rangle_{G_n}=u^\top G_n v$, the metric-normal direction is $G_n^{-1}r_n$, and the normal-distance denominator is $r_n^\top G_n^{-1}r_n$. Thus a geometry choice must supply a matched pair: $G_n$ acts on primal output displacement and $G_n^{-1}$ on the loss covector. An arbitrary denominator does not constitute a new geometry.

The canonical choice $G_n=I$ recovers iEF \citep{IEF}. More generally, any smooth output discrepancy whose second-order expansion supplies a valid local quadratic metric can enter IEWC through its matched primal/dual pair $G_n,G_n^{-1}$.

For ordered probability outputs, Appendix~\ref{app:ordered-cdf-geometry} constructs a CDF/Cram\'er metric and its matched pseudoinverse dual on the simplex tangent. This gives one explicit space-dependent output geometry. The diffusion experiment in Section~\ref{sec:empirical} uses a related cumulative-distribution output representation and shows that IEWC accommodates such alternatives.

\section{Improved Elastic Weight Consolidation}

The preceding derivation leaves five implementation choices: global penalty strength, damping, aggregation weights, output geometry, and matrix approximation. For old-task samples $a_1,\dots,a_N$, IEWC uses
\[
M_{\mathrm{IEWC}}
=
\frac1N
\sum_{n=1}^N
\bar h_n
\frac{g_ng_n^\top}{r_n^\top G_n^{-1}r_n+\tau_A},
\qquad
g_n=J_n^\top r_n,
\]
where $\bar h_n\ge0$ has empirical mean one. The continual-learning penalty is
\[
\frac{\lambda}{2}
(\theta-\theta_A^\star)^\top
M_{\mathrm{IEWC}}
(\theta-\theta_A^\star).
\]

The canonical geometry is $G_n=I$. We use task-relative damping $\tau_A=\kappa\operatorname{median}\{r_i^\top G_i^{-1}r_i:r_i\ne0\}$ with $\kappa=10^{-2}$ in the held-out evaluation. This makes $\kappa$ dimensionless and preserves global loss-unit invariance. If all denominators are zero, the stored importance is zero.

Uniform IEWC sets $\bar h_n=1$. To study weighting without confounding it with normalization, we define a separate \emph{residual-diversity weighting} (RDW) variant. RDW assigns mean-one weights from the component of each unit parameter-gradient direction outside a capped span of earlier directions; Appendix~\ref{app:rdw-rule} gives the exact streaming rule. Both OPG-RDW and IEF-RDW receive the same unit parameter-gradient directions. Scalar IEF normalization cannot change those directions, so the $2\times2$ comparison
\[
\{\mathrm{OPG},\mathrm{IEF}\}
\times
\{\mathrm{uniform},\mathrm{RDW}\}
\]
isolates matrix normalization from sample weighting.

An IEWC instantiation is specified by the five-tuple $(\lambda,\kappa,\Gamma,h,\Pi)$, where $\lambda$ sets preservation strength, $\kappa$ sets relative damping, $\Gamma$ maps an old output to its local metric $G_n$, $h$ supplies aggregation weights, and $\Pi$ is the matrix approximation operator (for example, full, diagonal, or low rank). These are design choices that the derivation leaves open. Scalar loss preservation gives OPG-EWC; loss-level-set normal preservation gives uniform IEWC. We call this CLOC-derived mechanism \emph{Improved Elastic Weight Consolidation} while treating its individual axes separately in theory and experiment.

\section{Empirical Evaluation}
\label{sec:empirical}

We report two evaluations. The multi-domain study measures IEWC's performance across classification, NLP, forecasting, segmentation, and diffusion and reports the best-performing configuration in each setting. The held-out evaluation fixes hyperparameter selection before any test access and estimates paired effects on held-out data. It compares IEF with OPG under matched uniform weighting and diagonal scale, then crosses OPG/IEF with uniform/RDW aggregation. Appendix C specifies data splits, fixed validation grids, repetitions, baselines, and uncertainty calculations.

Unless stated otherwise, all stored matrices are diagonal. EWC, IEWC, and IEWC+RDW denote observed-label OPG, uniform IEF, and residual-diversity-weighted IEF, respectively. RDW is order dependent, so the held-out evaluation separates its contribution from that of normalization; Appendix C maps these names to the archived code and gives complete protocols.

\subsection{Multi-domain results}

\begin{table}[t]
\caption[Multi-domain continual-learning results]{Multi-domain results, means over seeds of the selected configurations. Across-seed standard deviations and full-precision means are reported in Appendix Table~\ref{tab:main-empirical-results-uncertainty}. Higher is better except for MSE and forgetting.}
\label{tab:main-empirical-results}
\centering
\small
\setlength{\tabcolsep}{0pt}
\begin{tabular*}{\linewidth}{@{\extracolsep{\fill}}llllrrrrr@{}}
\toprule
Task & Benchmark & Model & Metric & Seq. & EWC (OPG) & EWC-DR & IEWC & IEWC+RDW \\
\midrule
\multirow{2}{*}{Classification} & \multirow{2}{*}{ImageNet-R} & \multirow{2}{*}{ViT-B/16} & Acc. $\uparrow$ & .2896 & .5981 & .5687 & \textbf{.6332} & .6250 \\
& & & Fgt. $\downarrow$ & .7066 & .3595 & .3933 & \textbf{.3142} & .3285 \\
\addlinespace[2pt]
\multirow{2}{*}{Language} & \multirow{2}{*}{TRACE} & \multirow{2}{*}{Qwen2.5-3B} & Score $\uparrow$ & .5760 & .5972 & .5972 & .5938 & \textbf{.6272} \\
& & & Fgt. $\downarrow$ & .1139 & .0762 & .0781 & .0820 & \textbf{.0046} \\
\addlinespace[2pt]
\multirow{2}{*}{Forecasting} & \multirow{2}{*}{Weather, ECL, Traffic} & \multirow{2}{*}{PatchTST} & MSE $\downarrow$ & .2733 & .2388 & -- & .2418 & \textbf{.2383} \\
& & & Fgt. $\downarrow$ & .0535 & \textbf{.0140} & -- & .0191 & .0147 \\
\addlinespace[2pt]
\multirow{2}{*}{Segmentation} & \multirow{2}{*}{VOC 15--5} & \multirow{2}{*}{ConvNeXtV2-T FPN} & mIoU $\uparrow$ & .0917 & .0946 & .3080 & \textbf{.4678} & .4274 \\
& & & Fgt. $\downarrow$ & .7135 & .7131 & .4471 & \textbf{.2418} & .2925 \\
\addlinespace[2pt]
\multirow{2}{*}{Diffusion} & \multirow{2}{*}{MNIST} & \multirow{2}{*}{DDPM} & MSE $\downarrow$ & .0676 & .0296 & -- & .0298 & \textbf{.0293} \\
& & & Fgt. $\downarrow$ & .0503 & \textbf{-.0006} & -- & .0001 & -.0003 \\
\bottomrule
\end{tabular*}
\end{table}

Table \ref{tab:main-empirical-results} shows that uniform IEWC achieves higher means than OPG-EWC on ImageNet-R and VOC 15--5, while weighted IEWC achieves the highest TRACE score and the lowest forecasting and DDPM MSE. The held-out evaluation then tests which gains generalize and whether they arise from normalization or weighting.

\begin{figure}[t]
\centering
\begin{tikzpicture}
\begin{groupplot}[group style={group size=2 by 2,horizontal sep=22mm,vertical sep=19mm},iewc axis,width=.37\linewidth,height=.19\linewidth]
\nextgroupplot[title={\textbf{A}\quad ImageNet-R performance},ylabel={Final accuracy},xmin=.6,xmax=10.4,xtick={1,...,10},xlabel={ImageNet-R distribution},legend to name=retentionlegend,legend columns=5]
\addplot[draw=none,fill=IEWCLight,fill opacity=.08,forget plot] coordinates {(1,0.0058824897) (2,0.054450644) (3,0.25996699) (4,0.28911963) (5,0.35766973) (6,0.37274447) (7,0.40776634) (8,0.32034382) (9,0.37695905) (10,0.93040358) (10,0.91814561) (9,0.21815736) (8,0.2037652) (7,0.32353228) (6,0.26856562) (5,0.20577113) (4,0.1112691) (3,0.15835731) (2,0.0079329683) (1,0.00025624574)} -- cycle;
\addplot+[sequential] coordinates {(1,0.0030693677) (2,0.031191806) (3,0.20916215) (4,0.20019436) (5,0.28172043) (6,0.32065504) (7,0.36564931) (8,0.26205451) (9,0.29755821) (10,0.92427459)};
\addlegendentry{Sequential}
\addplot[draw=none,fill=IEWCDark,fill opacity=.08,forget plot] coordinates {(1,0.3625675) (2,0.48798311) (3,0.69122021) (4,0.56991455) (5,0.58767422) (6,0.65753547) (7,0.62153356) (8,0.76003713) (9,0.74225433) (10,0.932826) (10,0.90723061) (9,0.68534362) (8,0.64876791) (7,0.61604873) (6,0.56697589) (5,0.50049783) (4,0.4777045) (3,0.57339703) (2,0.30438188) (1,0.26849451)} -- cycle;
\addplot+[ewc] coordinates {(1,0.315531) (2,0.3961825) (3,0.63230862) (4,0.52380952) (5,0.54408602) (6,0.61225568) (7,0.61879114) (8,0.70440252) (9,0.71379898) (10,0.92002831)};
\addlegendentry{EWC}
\addplot[draw=none,fill=IEWCMid,fill opacity=.08,forget plot] coordinates {(1,0.39979564) (2,0.31749301) (3,0.63054665) (4,0.52734172) (5,0.57063309) (6,0.61598993) (7,0.63944345) (8,0.68968755) (9,0.68939239) (10,0.92455223) (10,0.90984267) (9,0.6041908) (8,0.64154934) (7,0.59454818) (6,0.5239995) (5,0.4949583) (4,0.45322193) (3,0.53642141) (2,0.29889433) (1,0.31229767)} -- cycle;
\addplot+[ewcdr] coordinates {(1,0.35604665) (2,0.30819367) (3,0.58348403) (4,0.49028183) (5,0.5327957) (6,0.56999472) (7,0.61699581) (8,0.66561845) (9,0.6467916) (10,0.91719745)};
\addlegendentry{EWC-DR}
\addplot[draw=none,fill=IEWCPrimary,fill opacity=.08,forget plot] coordinates {(1,0.53840511) (2,0.5141191) (3,0.7006819) (4,0.57656183) (5,0.58781199) (6,0.64575189) (7,0.65063476) (8,0.73287238) (9,0.74134961) (10,0.90416318) (10,0.89767687) (9,0.71577702) (8,0.71471671) (7,0.6192635) (6,0.59249428) (5,0.56810199) (4,0.52547899) (3,0.64591244) (2,0.39183993) (1,0.39959366)} -- cycle;
\addplot+[iewc] coordinates {(1,0.46899939) (2,0.45297952) (3,0.67329717) (4,0.55102041) (5,0.57795699) (6,0.61912309) (7,0.63494913) (8,0.72379455) (9,0.72856332) (10,0.90092003)};
\addlegendentry{IEWC}
\addplot[draw=none,fill=IEWCSecondary,fill opacity=.08,forget plot] coordinates {(1,0.47758371) (2,0.45021498) (3,0.65869239) (4,0.61187664) (5,0.59271109) (6,0.66959395) (7,0.66860862) (8,0.77932388) (9,0.71489907) (10,0.91719745) (10,0.90021231) (9,0.70588458) (8,0.73744761) (7,0.66113405) (6,0.59612184) (5,0.53417063) (4,0.51543142) (3,0.61677476) (2,0.36728968) (1,0.32536288)} -- cycle;
\addplot+[iewcrdw] coordinates {(1,0.4014733) (2,0.40875233) (3,0.63773357) (4,0.56365403) (5,0.56344086) (6,0.6328579) (7,0.66487133) (8,0.75838574) (9,0.71039182) (10,0.90870488)};
\addlegendentry{IEWC+RDW}
\nextgroupplot[title={\textbf{B}\quad ImageNet-R forgetting},ylabel={Forgetting},xmin=.6,xmax=10.4,xtick={1,...,10},xlabel={ImageNet-R distribution}]
\addplot[draw=none,fill=IEWCLight,fill opacity=.08,forget plot] coordinates {(1,0.92734417) (2,0.88702302) (3,0.78695652) (4,0.79476292) (5,0.74285443) (6,0.66230484) (7,0.6085797) (8,0.75306599) (9,0.70100345) (10,0) (10,0) (9,0.54147241) (8,0.62743716) (7,0.51410133) (6,0.54318908) (5,0.59693051) (4,0.60465399) (3,0.670548) (2,0.84109616) (1,0.9155042)} -- cycle;
\addplot+[sequential] coordinates {(1,0.92142419) (2,0.86405959) (3,0.72875226) (4,0.69970845) (5,0.66989247) (6,0.60274696) (7,0.56134051) (8,0.69025157) (9,0.62123793) (10,0)};
\addplot[draw=none,fill=IEWCDark,fill opacity=.08,forget plot] coordinates {(1,0.65263395) (2,0.60007694) (3,0.36185296) (4,0.41250716) (5,0.44907281) (6,0.36048063) (7,0.31699778) (8,0.28180433) (9,0.22322142) (10,0) (10,0) (9,0.17428) (8,0.17626694) (7,0.29700581) (6,0.25864245) (5,0.35522827) (4,0.33190489) (3,0.2541808) (2,0.39899196) (1,0.56529116)} -- cycle;
\addplot+[ewc] coordinates {(1,0.60896255) (2,0.49953445) (3,0.30801688) (4,0.37220603) (5,0.40215054) (6,0.30956154) (7,0.3070018) (8,0.22903564) (9,0.19875071) (10,0)};
\addplot[draw=none,fill=IEWCMid,fill opacity=.08,forget plot] coordinates {(1,0.61036343) (2,0.60191475) (3,0.39760871) (4,0.44232301) (5,0.45210058) (6,0.39785792) (7,0.33579857) (8,0.29656036) (9,0.30846093) (10,0) (10,0) (9,0.23441244) (8,0.25480232) (7,0.2805988) (6,0.29944794) (5,0.37693168) (4,0.37108807) (3,0.31245759) (2,0.5805806) (1,0.52530262)} -- cycle;
\addplot+[ewcdr] coordinates {(1,0.56783303) (2,0.59124767) (3,0.35503315) (4,0.40670554) (5,0.41451613) (6,0.34865293) (7,0.30819868) (8,0.27568134) (9,0.27143668) (10,0)};
\addplot[draw=none,fill=IEWCPrimary,fill opacity=.08,forget plot] coordinates {(1,0.52302026) (2,0.51098123) (3,0.28055857) (4,0.37393472) (5,0.38199565) (6,0.31486067) (7,0.2998883) (8,0.22412579) (9,0.18119063) (10,0) (10,0) (9,0.16633918) (8,0.19725786) (7,0.2734211) (6,0.26411451) (5,0.34811188) (4,0.31702737) (3,0.24385372) (2,0.36890704) (1,0.38674033)} -- cycle;
\addplot+[iewc] coordinates {(1,0.45488029) (2,0.43994413) (3,0.26220615) (4,0.34548105) (5,0.36505376) (6,0.28948759) (7,0.2866547) (8,0.21069182) (9,0.17376491) (10,0)};
\addplot[draw=none,fill=IEWCSecondary,fill opacity=.08,forget plot] coordinates {(1,0.59647761) (2,0.53810183) (3,0.31146211) (4,0.3865027) (5,0.41782584) (6,0.3260245) (7,0.27303907) (8,0.20414498) (9,0.20949056) (10,0) (10,0) (9,0.18346798) (8,0.15853846) (7,0.25119791) (6,0.24238543) (5,0.34884083) (4,0.29085396) (3,0.28407738) (2,0.4404829) (1,0.44956291)} -- cycle;
\addplot+[iewcrdw] coordinates {(1,0.52302026) (2,0.48929236) (3,0.29776974) (4,0.33867833) (5,0.38333333) (6,0.28420497) (7,0.26211849) (8,0.18134172) (9,0.19647927) (10,0)};
\nextgroupplot[title={\textbf{C}\quad TRACE performance},ylabel={Final score},xmin=.75,xmax=3.25,xtick={1,2,3},xticklabels={C-STANCE,FOMC,ScienceQA},xlabel={TRACE task}]
\addplot[draw=none,fill=IEWCLight,fill opacity=.08,forget plot] coordinates {(1,0.436628) (2,0.54801368) (3,0.80031725) (3,0.77520358) (2,0.50667382) (1,0.38889283)} -- cycle;
\addplot+[sequential] coordinates {(1,0.41276042) (2,0.52734375) (3,0.78776042)};
\addplot[draw=none,fill=IEWCDark,fill opacity=.08,forget plot] coordinates {(1,0.48046875) (2,0.57649444) (3,0.80146842) (3,0.76623991) (2,0.54069306) (1,0.41796875)} -- cycle;
\addplot+[ewc] coordinates {(1,0.44921875) (2,0.55859375) (3,0.78385417)};
\addplot[draw=none,fill=IEWCMid,fill opacity=.08,forget plot] coordinates {(1,0.47818843) (2,0.60017051) (3,0.79510892) (3,0.76999525) (2,0.52482949) (1,0.41504074)} -- cycle;
\addplot+[ewcdr] coordinates {(1,0.44661458) (2,0.5625) (3,0.78255208)};
\addplot[draw=none,fill=IEWCPrimary,fill opacity=.08,forget plot] coordinates {(1,0.47241115) (2,0.59375) (3,0.7943707) (3,0.7603168) (2,0.53125) (1,0.41040135)} -- cycle;
\addplot+[iewc] coordinates {(1,0.44140625) (2,0.5625) (3,0.77734375)};
\addplot[draw=none,fill=IEWCSecondary,fill opacity=.08,forget plot] coordinates {(1,0.578125) (2,0.59260984) (3,0.75642872) (3,0.73575878) (2,0.561036) (1,0.5390625)} -- cycle;
\addplot+[iewcrdw] coordinates {(1,0.55859375) (2,0.57682292) (3,0.74609375)};
\nextgroupplot[title={\textbf{D}\quad TRACE forgetting},ylabel={Forgetting},xmin=.75,xmax=3.25,xtick={1,2,3},xticklabels={C-STANCE,FOMC,ScienceQA},xlabel={TRACE task}]
\addplot[draw=none,fill=IEWCLight,fill opacity=.08,forget plot] coordinates {(1,0.15234375) (2,0.10693275) (3,0) (3,0) (2,0.075358912) (1,0.12109375)} -- cycle;
\addplot+[sequential] coordinates {(1,0.13671875) (2,0.091145833) (3,0)};
\addplot[draw=none,fill=IEWCDark,fill opacity=.08,forget plot] coordinates {(1,0.10407318) (2,0.065423174) (3,0) (3,0) (2,0.049160159) (1,0.086030985)} -- cycle;
\addplot+[ewc] coordinates {(1,0.095052083) (2,0.057291667) (3,0)};
\addplot[draw=none,fill=IEWCMid,fill opacity=.08,forget plot] coordinates {(1,0.11479837) (2,0.085199978) (3,0) (3,0) (2,0.021570855) (1,0.090930792)} -- cycle;
\addplot+[ewcdr] coordinates {(1,0.10286458) (2,0.053385417) (3,0)};
\addplot[draw=none,fill=IEWCPrimary,fill opacity=.08,forget plot] coordinates {(1,0.11580372) (2,0.091968765) (3,0) (3,0) (2,0.025218735) (1,0.095133784)} -- cycle;
\addplot+[iewc] coordinates {(1,0.10546875) (2,0.05859375) (3,0)};
\addplot[draw=none,fill=IEWCSecondary,fill opacity=.08,forget plot] coordinates {(1,0) (2,0.024901505) (3,0.026297073) (3,0.012765427) (2,-0.0066723381) (1,0)} -- cycle;
\addplot+[iewcrdw] coordinates {(1,0) (2,0.0091145833) (3,0.01953125)};
\end{groupplot}
\node[anchor=south] at ($(group c1r1.north west)!0.5!(group c2r1.north east)+(0,7mm)$) {\pgfplotslegendfromname{retentionlegend}};
\end{tikzpicture}
\caption[ImageNet-R and TRACE distribution retention]{Distribution-level retention in the multi-domain study. \textbf{A--B}: ImageNet-R final accuracy and forgetting. \textbf{C--D}: TRACE final score and forgetting for C-STANCE, FOMC, and ScienceQA. Bands show across-seed standard deviations.}
\label{fig:distribution-retention}
\end{figure}
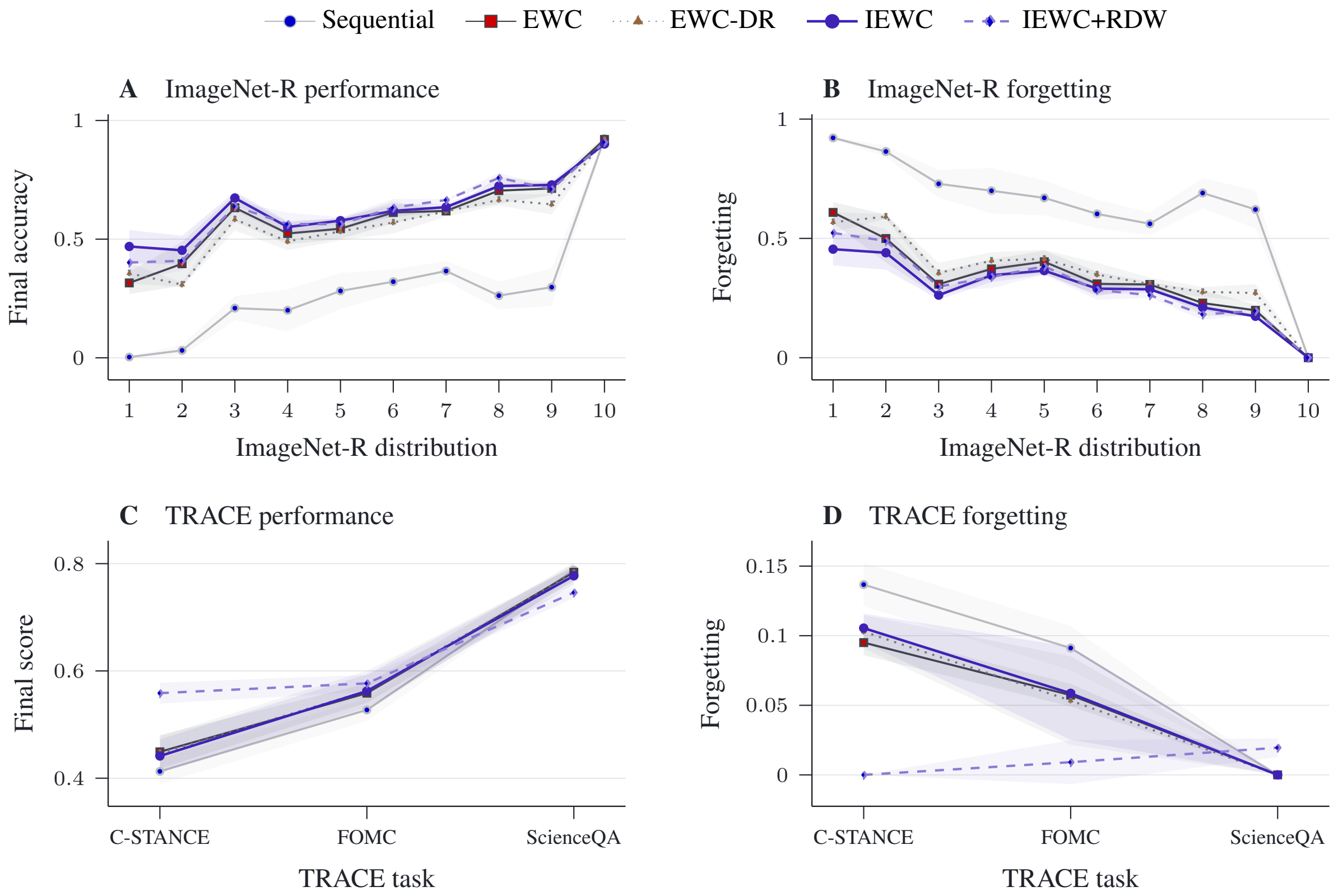

At the distribution level, Figure~\ref{fig:distribution-retention}A--B shows that IEWC's retained accuracy is not concentrated in a single late distribution. The held-out evaluation tests whether this advantage generalizes across fixed class orders. The weighted IEWC run likewise has less forgetting on C-STANCE and FOMC in Figure~\ref{fig:distribution-retention}C--D. The held-out evaluation crosses the weighting mechanism with both OPG and IEF, and replaces FOMC with NumGLUE-cm to obtain disjoint validation and test splits.

Weighted IEWC retains more target-trajectory structure at the first task and remains close to OPG-EWC later; Appendix Figure~\ref{fig:forecast-rollouts} provides representative held-out rollouts alongside the aggregate scores.

The DDPM study also uses a row/column cumulative-distribution output representation. The selected run reduces old-output drift to $9.4409\pm7.2517$, compared with $14.1026\pm12.2220$ for Euclidean uniform IEWC, while attaining denoising MSE $0.0301\pm0.0012$ versus $0.0298\pm0.0014$. IEWC thus accommodates alternative output representations, and Appendix~\ref{app:ordered-cdf-geometry} gives an ordered-CDF/Cram\'er geometry with a matched primal/dual pair.

Section~\ref{sec:normalization-weighting} predicts that, where output-loss-gradient scales vary across samples, removing gradient scale narrows the spread of per-sample importance. Figure~\ref{fig:locked-summands}C--F shows the effective per-sample diagonal summands, $\lambda$ times the per-sample diagonal trace before averaging, in four representative settings. In classification and regression, OPG summands span more than three orders of magnitude between their 5th and 95th percentiles, while IEWC narrows this range by more than an order of magnitude; in diffusion and segmentation, both spreads are comparable.

\subsection{Held-out evaluation}

\begin{figure}[t]
\centering
\begin{tikzpicture}
\begin{groupplot}[group style={group size=2 by 1,horizontal sep=22mm},iewc axis,width=.37\linewidth,height=.09\linewidth,xmin=-5.5,xmax=5.5,ymin=-1,ymax=1,ytick=\empty,axis y line=none,ymajorgrids=false,xmajorgrids,xlabel={IEF $-$ OPG, uniform (percentage points)}]
\nextgroupplot[title={\textbf{A}\quad ImageNet-R ($n=10$)}]
\addplot[draw=none,fill=IEWCBand,forget plot] coordinates {(-1,-1)(1,-1)(1,1)(-1,1)} -- cycle;
\addplot[IEWCMid,line width=.45pt,forget plot] coordinates {(0,-1)(0,1)};
\addplot[only marks,mark=*,mark size=1.15pt,color=IEWCMid] coordinates {(2.3931215,-0.55) (2.4719204,-0.42777778) (3.707981,-0.30555556) (4.1576971,-0.18333333) (2.1177626,-0.061111111) (3.1135012,0.061111111) (-0.16151123,0.18333333) (1.5992166,0.30555556) (0.63030049,0.42777778) (-0.027104204,0.55)};
\addplot[IEWCPrimary,line width=.85pt] coordinates {(0.93331513,0)(3.0672619,0)};
\addplot[IEWCPrimary,line width=.85pt] coordinates {(0.93331513,-.18)(0.93331513,.18)};
\addplot[IEWCPrimary,line width=.85pt] coordinates {(3.0672619,-.18)(3.0672619,.18)};
\addplot[only marks,mark=diamond*,mark size=2.4pt,color=IEWCPrimary] coordinates {(2.0002885,0)};
\nextgroupplot[title={\textbf{B}\quad TRACE ($n=12$)}]
\addplot[draw=none,fill=IEWCBand,forget plot] coordinates {(-1,-1)(1,-1)(1,1)(-1,1)} -- cycle;
\addplot[IEWCMid,line width=.45pt,forget plot] coordinates {(0,-1)(0,1)};
\addplot[only marks,mark=*,mark size=1.15pt,color=IEWCMid] coordinates {(0.4781893,-0.55) (1.9833333,-0.45) (-0.83971193,-0.35) (-2.1076132,-0.25) (-0.0781893,-0.15) (-5.0676955,-0.05) (0.27695473,0.05) (3.987037,0.15) (0.74485597,0.25) (-0.05,0.35) (-0.44485597,0.45) (-0.16666667,0.55)};
\addplot[IEWCPrimary,line width=.85pt] coordinates {(-1.4833562,0)(1.2692959,0)};
\addplot[IEWCPrimary,line width=.85pt] coordinates {(-1.4833562,-.18)(-1.4833562,.18)};
\addplot[IEWCPrimary,line width=.85pt] coordinates {(1.2692959,-.18)(1.2692959,.18)};
\addplot[only marks,mark=diamond*,mark size=2.4pt,color=IEWCPrimary] coordinates {(-0.10703018,0)};
\end{groupplot}
\end{tikzpicture}

\vspace{2mm}
\input{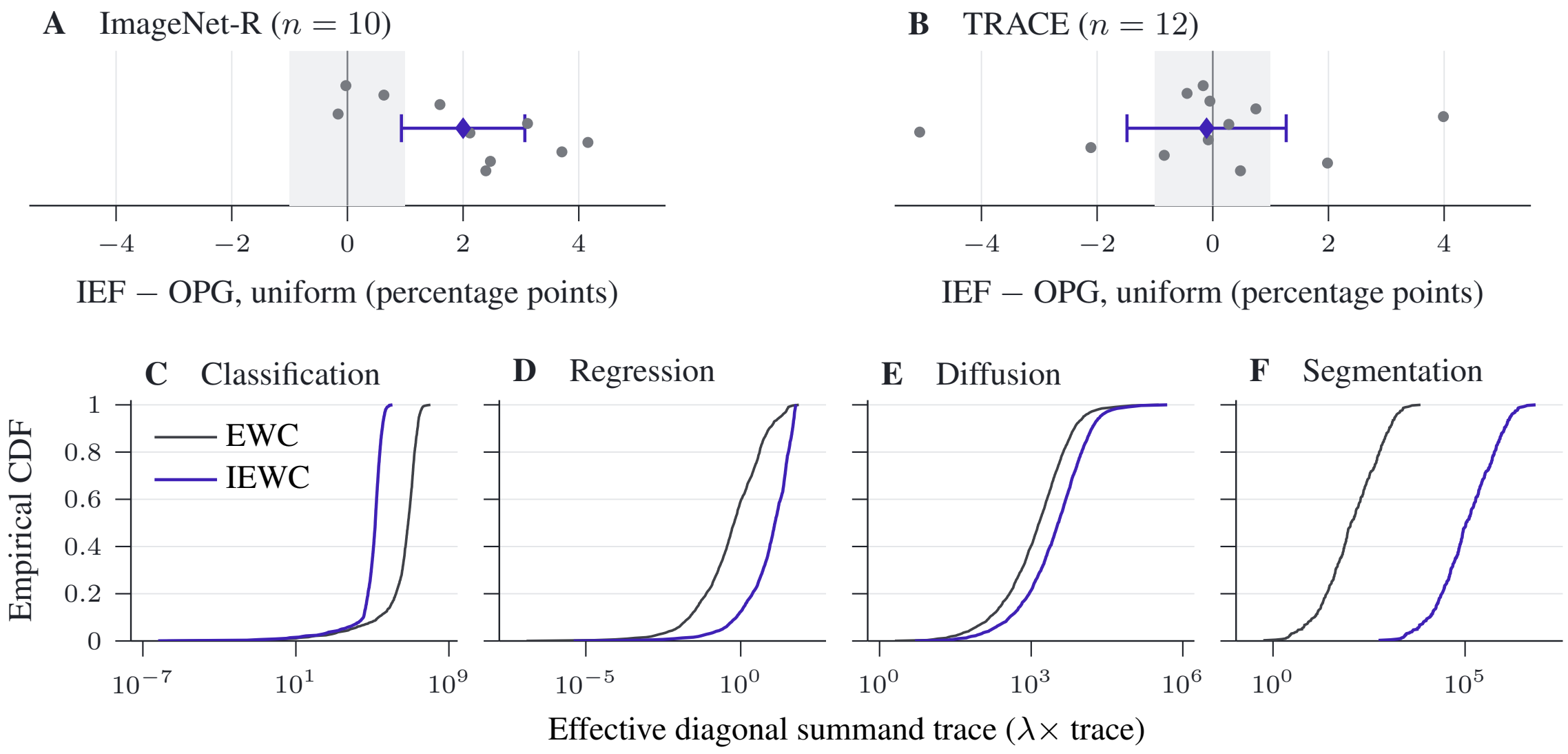}
\caption[Held-out paired effects and effective summand norms]{\textbf{A--B}: Held-out paired differences in final average performance, uniform IEF minus uniform OPG. Gray points are individual paired runs, the diamond is the mean, the bar its 95\% Student-$t$ interval, and the shaded band the one-point practical threshold fixed in advance. \textbf{C--F}: Empirical CDFs of $\lambda$ times the per-sample diagonal trace for CIFAR-100 classification, controlled regression, MNIST DDPM denoising, and binary segmentation.}
\label{fig:locked-summands}
\end{figure}

The held-out evaluation selects $\lambda$ on validation data from fixed five-point grids and evaluates each method once on the test split (Appendix~\ref{app:confirmatory-results}). ImageNet-R uses ten paired class orders; TRACE crosses all six orders of C-STANCE, ScienceQA, and NumGLUE-cm with two seeds. On ImageNet-R, uniform IEWC improves final average accuracy over matched EWC by 2.00 percentage points (95\% CI 0.93--3.07; exact sign-flip $p=0.0078$; Figure~\ref{fig:locked-summands}A), and the IEF advantage persists under RDW aggregation (1.71 points, 0.92--2.51; Holm $p=0.016$). On TRACE, the two estimators perform comparably (Figure~\ref{fig:locked-summands}B).

\section{Limitations}

CLOC is conditional on its preservation specification: it derives a penalty after the preserved functionals and constraint sets have been chosen, but it does not determine which properties are sufficient to preserve old-task competence under future updates. Misspecified or mutually incompatible specifications can therefore protect the wrong behavior even when their local penalties are optimized accurately. Selecting or learning suitable preservation specifications is a natural next step.

IEWC, like EWC, compresses an old task into one importance matrix evaluated at $\theta_A^\star$ and keeps it fixed during later training. Before aggregation, however, its normalized parameter-sensitivity direction $a^{G}(x;\theta_A^\star)=J_{\theta_A^\star}(x)^\top\hat r^{G}(x)$ is input-dependent. Circuit analyses make the same distinction between a model's fixed connections and the input-dependent feature contributions that realize a particular computation \citep{DunefskyTranscoders}. A single average, especially after diagonalization, can therefore hide sensitivity confined to subsets of inputs or distributed across overlapping pathways, and cannot track changes in the relevant Jacobians as the parameters move away from $\theta_A^\star$. Sample weighting changes which empirical average is stored, but cannot recover this conditional structure after compression. Conditional mixtures or dynamically refreshed importance are natural extensions, at the cost of richer stored summaries or access to old inputs.

Finally, the empirical comparisons cover selected datasets, model families, adaptation regimes, and task orders, and hold the stored penalty form fixed across methods. Replay, distillation, and parameter-isolation methods make different storage or architectural assumptions and lie outside this comparison.

\section{Conclusion}

CLOC is a preservation-first formulation deriving local penalties from selected learned-behavior functionals. Scalar-loss preservation yields an OPG-based regularization that under NLL equals EWC's empirical-Fisher regularization, whereas preserving outputs modulo old-loss level-set tangents yields IEWC and explains its normalization geometrically. CLOC also makes damping, output geometry, aggregation, and matrix approximation explicit design choices. Mechanism tests, the multi-domain study, and the held-out evaluation show that IEWC variants match or improve on EWC, including a 2.00-point gain in held-out ImageNet-R accuracy. Choosing what to preserve, rather than which curvature to approximate, is thus the decision from which a continual-learning penalty follows.

\subsection*{AI Use Statement}

The original IEWC framework and initial manuscript predate the generative-AI
assistance disclosed here. Generative AI tools were used to assist with revision
planning; theoretical and experimental development; implementation and
experiment execution; analysis and interpretation; figure creation; literature
search and synthesis; and manuscript drafting and editing. The author reviewed
AI-assisted code, experimental outputs, numerical claims, citations, figures,
and text against saved artifacts and executable code, and takes responsibility
for the final content.

\bibliographystyle{apalike}
\bibliography{references}

\ifarxiv\else
\input{sections/sec17_reproducibility_checklist.tex}
\fi

\clearpage
\appendix
\ifarxiv\else
\onecolumn
\aistatstitle{Improved Elastic Weight Consolidation as an Optimization Constraint for Continual Learning: Supplementary Materials}
\fi

\section{Invariance to Loss Reparameterization}

This appendix treats the undamped case $\tau=0$.

The output-level feasible set

\[
\mathcal C_n=\{z:\ell_n(z)=\ell_n(z_n^A)\}
\]

depends on the level sets of $\ell_n$, not on the numerical labels assigned to those level sets. If we apply a local monotone scalar transformation

\[
\tilde \ell=\phi(\ell),
\qquad
\phi'(\ell)>0,
\]

then the output gradient rescales as

\[
\tilde r_n
=
\phi'(\ell_n(z_n^A))\,r_n.
\]

For OPG, the sample contribution changes by

\[
J_n^\top \tilde r_n\tilde r_n^\top J_n
=
\phi'(\ell_n(z_n^A))^2
J_n^\top r_nr_n^\top J_n.
\]

For IEF, the same scalar factor appears in numerator and denominator:

\[
\frac{\tilde r_n\tilde r_n^\top}
{\tilde r_n^\top G_n^{-1}\tilde r_n}
=
\frac{r_nr_n^\top}
{r_n^\top G_n^{-1}r_n}.
\]

Thus undamped IEF is invariant to local scalar relabelings of the constraint function wherever $r_n\ne0$. This includes loss rescaling, unit changes in regression losses, reduction conventions, and smooth monotone transformations that preserve the local level-set geometry. OPG's global scale can be absorbed into $\lambda$ only when the transformation rescales all samples equally; sample-dependent factors change its relative weighting.

Positive shared damping narrows the invariance. A fixed numerical $\tau$ does not transform with the loss and therefore changes even under a global unit rescaling. Task-relative damping $\tau_A=\kappa\operatorname{median}_{i:s_i>0}s_i$ restores invariance to a global task-loss multiplier because $s_i$ and $\tau_A$ scale together. It does not generally preserve invariance to sample-dependent monotone transformations. At $r_n=0$, the implementation assigns a zero contribution rather than claiming an undamped $0/0$ identity.

\section{Contamination Contributions and Spectral Stability}
\label{app:contamination}

The OPG--IEF normalization difference can change which samples dominate a full importance matrix. In Euclidean output geometry, write $s_i=\|r_i\|_2^2$ and $a_i=g_i/\sqrt{s_i}$ for $s_i>0$. The undamped per-sample matrices are
\[
M_{\mathrm{OPG},i}=s_i a_i a_i^\top,
\qquad
M_{\mathrm{IEF},i}=a_i a_i^\top.
\]
Thus OPG couples each sensitivity direction to the squared output-loss-gradient scale, whereas undamped IEF keeps the direction-only outer product. Relative damping multiplies IEF by $s_i/(s_i+\tau_A)$ and therefore retains the near-zero qualification from Section~\ref{sec:well-fitted}.

For intuition, suppose a fraction $\varepsilon$ of samples are contaminated, their normalized sensitivity is concentrated near a direction $u$ outside the leading clean subspace, and their scale is $s_i\approx W$. If clean scales are sufficiently concentrated or independent of direction, then
\[
\mathbb E[M_{\mathrm{OPG}}]
\approx
(1-\varepsilon)\bar s_{\mathrm{clean}}\Sigma_{\mathrm{clean}}
+\varepsilon Wuu^\top,
\]
whereas undamped IEF has no factor $W$ on the contaminating direction. In particular, a sufficiently large $\varepsilon W$ can move $u$ into the leading OPG eigenspace. This model is an illustration; the experiment below tests the mechanism directly.

We test the mechanism directly in a fitted two-class MLP. For each seed, 512 inputs are drawn from $\mathcal N(0,I_2)$ and assigned the clean label $\mathbf{1}\{x_1+0.75x_2<0\}$; a uniformly selected $0$, $5$, $10$, or $20\%$ of labels is flipped. A $2$--$32$--$2$ tanh MLP is trained for 250 epochs with Adam, learning rate $0.01$, and batch size 64. For each of 20 independent seeds we then retain every per-sample parameter-gradient row and clean/corrupt identity. We decompose the full OPG and task-relative-damped IEF matrices ($\kappa=0.01$) into clean and corrupt contributions, then compare each total matrix with its clean-contribution reference. Table~\ref{tab:contamination-mechanism} reports means over seeds.

\begin{table}[H]
\caption{Direct full-matrix contamination diagnostic. Alignment is the absolute inner product of the leading eigenvectors; top-5 overlap is the mean squared singular value between the corresponding eigenspaces. Each reference is the clean contribution from the same fitted model.}
\label{tab:contamination-mechanism}
\centering
\small
\resizebox{\linewidth}{!}{%
\begin{tabular}{rrrrrrrr}
\toprule
Flip rate & Corrupt/clean $s_i$ & \multicolumn{2}{c}{Corrupt trace share} & \multicolumn{2}{c}{Leading-vector alignment} & \multicolumn{2}{c}{Top-5 overlap} \\
& ratio & OPG & IEF & OPG & IEF & OPG & IEF \\
\midrule
$0.05$ & 81.26 & 0.783 & 0.054 & 0.426 & 1.000 & 0.902 & 0.996 \\
$0.10$ & 36.67 & 0.789 & 0.104 & 0.720 & 0.999 & 0.923 & 0.997 \\
$0.20$ & 11.23 & 0.736 & 0.202 & 0.875 & 0.998 & 0.930 & 0.990 \\
\bottomrule
\end{tabular}}
\end{table}

\begin{table}[H]
\caption{Paired mean differences with 95\% Student-$t$ intervals over 20 seeds. Every interval excludes zero.}
\label{tab:contamination-mechanism-intervals}
\centering
\small
\begin{tabular}{lccc}
\toprule
IEF minus OPG & 5\% flips & 10\% flips & 20\% flips \\
\midrule
Corrupt trace share & $-0.729$ [$-0.751,-0.706$] & $-0.685$ [$-0.701,-0.670$] & $-0.534$ [$-0.549,-0.519$] \\
Leading-vector alignment & $+0.574$ [$+0.444,+0.704$] & $+0.279$ [$+0.152,+0.406$] & $+0.123$ [$+0.034,+0.213$] \\
Top-5 overlap & $+0.094$ [$+0.067,+0.122$] & $+0.074$ [$+0.047,+0.101$] & $+0.060$ [$+0.029,+0.091$] \\
\bottomrule
\end{tabular}
\end{table}

These observations support the proposed contamination mechanism in this controlled construction; Appendix~\ref{app:label-noise} gives the downstream benchmark view.

\section{Experimental Protocol, Provenance, and Sensitivity}

\subsection{Held-out evaluation protocol}

The evaluation protocol was fixed and hashed before any test result was computed. It specifies the data splits, methods, hyperparameter grids, repetitions, metrics, and statistical analysis. Test data can be loaded only with the matching protocol hash, and every access is logged before examples are loaded. Validation scripts refuse to load test data, and validation runs do not enter the reported intervals.

The primary comparison in each domain is the paired final average performance of uniform IEF minus uniform OPG. The secondary comparisons cross estimator and aggregation:
\[
\{\mathrm{OPG},\mathrm{IEF}\}
\times
\{\mathrm{uniform},\mathrm{RDW}\}.
\]
All four combinations use the same importance examples, and the RDW variants receive the same unit-normalized parameter-gradient directions. Sampled model-Fisher EWC and MAS \citep{MAS} are additional regularization baselines. Sequential and offline/joint training are references. Every nonzero diagonal importance is rescaled to mean one across trainable parameters, and every method uses the displayed $\lambda/2$ convention.

This comparison keeps the same post-task diagonal penalty for every method. SI's trajectory accumulation and LwF's distillation objective are discussed in Section~\ref{sec:related-work} but are not part of it. EWC, IEWC, and IEWC+RDW in the paper correspond to the EF-EWC, IEWC-E/U-IEWC, and IEWC-GSS/GSS-IEWC labels in the archived code.

\subsubsection{Residual-diversity weighting rule}
\label{app:rdw-rule}

For each nonzero parameter gradient, let
\[
u_n=\frac{g_n}{\|g_n\|_2}.
\]
In a fixed sample order, RDW maintains an orthonormal basis $Q$ of at most $K$ earlier residual directions and assigns
\[
h_n=\|(I-Q^\top Q)u_n\|_2^2.
\]
When the residual is nonzero and the cap is not reached, its normalized direction is appended to $Q$; the final $h_n$ are mean-normalized. This streaming rule is motivated by residual-gradient diversity ideas in GSS \citep{GSS}, but it is an order-dependent aggregation heuristic, not part of the level-set derivation.

\begin{table}[H]
\caption{Data splits and repetitions in the held-out evaluation. FOMC is excluded because its distributed evaluation and test files are exact duplicates.}
\label{tab:confirmatory-protocol}
\centering
\small
\resizebox{\linewidth}{!}{%
\begin{tabular}{@{}p{0.11\linewidth}p{0.38\linewidth}p{0.18\linewidth}p{0.25\linewidth}@{}}
\toprule
Domain & Data splits & Validation runs & Test runs \\
\midrule
ImageNet-R & Stratified 80\%/20\% split of the original 24,000-example train set for train/test; the original 6,000-example test set is used only for validation & 3 class orders & 10 paired class orders \\
TRACE & Distributed train/eval/test files for C-STANCE, ScienceQA, and NumGLUE-cm; prompt/answer overlaps are zero & 3 task orders & 6 task orders $\times$ 2 optimizer seeds \\
\bottomrule
\end{tabular}}
\end{table}

ImageNet-R uses the fixed $\lambda$ grid $\{0.01,0.03,0.1,0.3,1\}$. TRACE uses $10^{-5}\!\times\!\{1,3,10,30,100\}$. Neither grid is extended after validation.

ImageNet-R uses a pretrained ViT-B/16 with LoRA rank 4 and ten 20-class tasks, trained for two epochs per task with AdamW at learning rate $3\cdot10^{-4}$ and batch size 64. Importance uses 128 examples per completed task and RDW cap 64. TRACE uses Qwen2.5-3B-Instruct with LoRA rank 8, two epochs, AdamW at $2\cdot10^{-4}$, batch size 2 with gradient accumulation 8, 512 training examples per task, and 64 importance examples per completed task. Data, loader, optimizer, task-order, importance-order, and sampled-Fisher-label seeds are separately recorded and paired where appropriate.

All runs use explicit DataLoader generators, reseed each task's training stage independently of the importance computation, and require deterministic PyTorch and CUDA kernels. Before validation, two complete runs with the same configuration were compared in each domain after removing wall-clock metadata; their training losses, evaluation matrices, and, for TRACE, generated example-level predictions matched exactly. These duplicate runs are kept for reproducibility checks but excluded from selection.

For every paired comparison we report the mean and standard deviation of paired differences, a two-sided 95\% Student-$t$ interval, the exact two-sided sign-flip $p$-value when enumerable, the median, and every paired value. These summaries treat task-order/seed pairs as independent; the interval assumes approximately normal paired differences, and the sign-flip test assumes that signs are exchangeable under the null. We interpret results against a practical threshold of $0.01$, one percentage point on either zero-to-one final-average metric, fixed in advance rather than relying on a binary significance rule. Holm correction is applied to all five secondary comparisons within a domain. Full five-point validation curves, importance time, total run time, and peak CUDA memory are recorded.

TRACE's twelve runs form a balanced cross of six task permutations and two model/optimizer seeds, so they are not twelve independent order draws. The primary analysis, threshold, and secondary comparisons therefore use all twelve runs, and two dependence checks complement them. First, we average each method over the two seeds within an order and recompute all paired summaries over the six orders. Second, we average over all six orders within each seed and report the two seed-level paired differences without treating a two-unit interval as reliable inference. The balanced design preserves the overall paired mean. Where run-level and order-level uncertainty differ, we base claims about TRACE on the more conservative order-level result. The six orders are the complete set of permutations, not a sample from a larger population of orders.

IEF uses task-relative damping
\[
\tau_A=10^{-2}\operatorname{median}\{s_i:s_i>0\}.
\]
Validation-only sensitivity varies the coefficient over $\{0,10^{-3},10^{-2},10^{-1},1\}$, importance counts, RDW caps, and RDW sample order. A separate loss-unit diagnostic multiplies a fitted classification loss by $0.1$, $1$, and $10$. OPG raw scale changes quadratically, fixed numerical damping changes IEF shape, and task-relative damping preserves the IEF matrix to numerical tolerance, matching Section~\ref{sec:well-fitted}.

\subsection{Held-out results}
\label{app:confirmatory-results}

Table~\ref{tab:confirmatory-main} reports every method in the held-out evaluation, and Table~\ref{tab:confirmatory-inference} the primary comparison and the Holm-corrected secondary comparisons. Values are fractions; differences in the main text are in percentage points.

\begin{table}[H]
\caption{Held-out final performance and forgetting, mean $\pm$ standard deviation over paired runs, with the validation-selected $\lambda$.}
\label{tab:confirmatory-main}
\centering
\small
\begin{tabular}{llccc}
\toprule
Domain & Method & Selected $\lambda$ & Final performance & Forgetting \\
\midrule
ImageNet-R & Sequential & -- & 0.2808 $\pm$ 0.0438 & 0.7018 $\pm$ 0.0467 \\
 & Offline/joint & -- & 0.8199 $\pm$ 0.0029 & 0.0000 $\pm$ 0.0000 \\
 & OPG, uniform & 1.0000 & 0.6416 $\pm$ 0.0301 & 0.2835 $\pm$ 0.0337 \\
 & IEF, uniform & 1.0000 & 0.6616 $\pm$ 0.0168 & 0.2566 $\pm$ 0.0177 \\
 & OPG, RDW & 1.0000 & 0.6445 $\pm$ 0.0278 & 0.2803 $\pm$ 0.0310 \\
 & IEF, RDW & 1.0000 & 0.6617 $\pm$ 0.0185 & 0.2565 $\pm$ 0.0196 \\
 & Sampled model Fisher & 1.0000 & 0.6477 $\pm$ 0.0242 & 0.2755 $\pm$ 0.0259 \\
 & MAS & 0.0300 & 0.4602 $\pm$ 0.0261 & 0.5015 $\pm$ 0.0273 \\
\midrule
TRACE & Sequential & -- & 0.5697 $\pm$ 0.0431 & 0.0758 $\pm$ 0.0440 \\
 & Offline/joint & -- & 0.6391 $\pm$ 0.0183 & 0.0000 $\pm$ 0.0000 \\
 & OPG, uniform & $3\times 10^{-4}$ & 0.5703 $\pm$ 0.0395 & 0.0750 $\pm$ 0.0458 \\
 & IEF, uniform & $3\times 10^{-4}$ & 0.5692 $\pm$ 0.0408 & 0.0670 $\pm$ 0.0471 \\
 & OPG, RDW & $3\times 10^{-4}$ & 0.5800 $\pm$ 0.0381 & 0.0685 $\pm$ 0.0446 \\
 & IEF, RDW & $3\times 10^{-5}$ & 0.5646 $\pm$ 0.0395 & 0.0763 $\pm$ 0.0407 \\
 & Sampled model Fisher & $1\times 10^{-5}$ & 0.5758 $\pm$ 0.0460 & 0.0759 $\pm$ 0.0427 \\
 & MAS & $1\times 10^{-4}$ & 0.5650 $\pm$ 0.0441 & 0.0785 $\pm$ 0.0433 \\
\bottomrule
\end{tabular}
\end{table}

\begin{table}[H]
\caption{Held-out paired comparisons with 95\% Student-$t$ intervals, exact sign-flip $p$-values, and Holm-corrected $p$-values for the secondary comparisons.}
\label{tab:confirmatory-inference}
\centering
\small
\resizebox{\linewidth}{!}{%
\begin{tabular}{lllcccc}
\toprule
Domain & Type & Comparison & Mean paired difference & 95\% CI & Exact $p$ & Holm $p$ \\
\midrule
ImageNet-R & Primary & IEF uniform $-$ OPG uniform & 0.0200 & [0.0093, 0.0307] & 0.0078 & -- \\
 & Secondary & IEF, RDW $-$ OPG, RDW & 0.0171 & [0.0092, 0.0251] & 0.0039 & 0.0156 \\
 & Secondary & OPG, RDW $-$ OPG, uniform & 0.0030 & [-0.0024, 0.0084] & 0.2637 & 0.5273 \\
 & Secondary & IEF, RDW $-$ IEF, uniform & $8.5\times 10^{-5}$ & [-0.0017, 0.0018] & 0.8867 & 0.8867 \\
 & Secondary & Sampled model Fisher $-$ OPG, uniform & 0.0061 & [-0.0021, 0.0144] & 0.1270 & 0.3809 \\
 & Secondary & MAS $-$ OPG, uniform & -0.1814 & [-0.1986, -0.1643] & 0.0020 & 0.0098 \\
\midrule
TRACE & Primary & IEF uniform $-$ OPG uniform & -0.0011 & [-0.0148, 0.0127] & 0.8589 & -- \\
 & Secondary & IEF, RDW $-$ OPG, RDW & -0.0155 & [-0.0335, 0.0026] & 0.0513 & 0.2563 \\
 & Secondary & OPG, RDW $-$ OPG, uniform & 0.0097 & [-0.0042, 0.0237] & 0.1914 & 0.7656 \\
 & Secondary & IEF, RDW $-$ IEF, uniform & -0.0047 & [-0.0136, 0.0043] & 0.2759 & 0.8276 \\
 & Secondary & Sampled model Fisher $-$ OPG, uniform & 0.0055 & [-0.0063, 0.0172] & 0.3887 & 0.8276 \\
 & Secondary & MAS $-$ OPG, uniform & -0.0053 & [-0.0156, 0.0050] & 0.3013 & 0.8276 \\
\bottomrule
\end{tabular}}
\end{table}

\subsection{Multi-domain protocol}

Table~\ref{tab:empirical-protocol} records the configurations behind Table~\ref{tab:main-empirical-results}. Its hyperparameters were selected per setting on the reported metrics; the held-out evaluation provides the validation-selected comparisons. Its implementation omits the factor $1/2$ of the displayed penalty, so its $\lambda$ values are not directly comparable with those of the held-out evaluation.

\begin{table}[H]
\caption{Complete multi-domain results corresponding to Table~\ref{tab:main-empirical-results}, reported as mean (across-seed standard deviation). Higher is better for accuracy, score, and mIoU; lower is better for MSE and forgetting.}
\label{tab:main-empirical-results-uncertainty}
\centering
\small
\setlength{\tabcolsep}{2.5pt}
\renewcommand{\arraystretch}{0.96}
\begin{tabular}{@{}llccccc@{}}
\toprule
Task & Metric & Seq. & EWC (OPG) & EWC-DR & IEWC & IEWC+RDW \\
\midrule
Classification & Final avg. acc. $\uparrow$ & 0.2896(.0405) & 0.5981(.0337) & 0.5687(.0263) & \textbf{0.6332(.0232)} & 0.6250(.0263) \\
& Avg. forgetting $\downarrow$ & 0.7066(.0466) & 0.3595(.0378) & 0.3933(.0286) & \textbf{0.3142(.0250)} & 0.3285(.0278) \\
\addlinespace[1pt]
NLP & Final avg. score $\uparrow$ & 0.5760(.0027) & 0.5972(.0020) & 0.5972(.0059) & 0.5938(.0034) & \textbf{0.6272(.0080)} \\
& Avg. forgetting $\downarrow$ & 0.1139(.0111) & 0.0762(.0034) & 0.0781(.0109) & 0.0820(.0122) & \textbf{0.0046(.0079)} \\
\addlinespace[1pt]
Forecasting & Final avg. MSE $\downarrow$ & 0.2733(.0065) & 0.2388(.0021) & -- & 0.2418(.0038) & \textbf{0.2383(.0007)} \\
& MSE forgetting $\downarrow$ & 0.0535(.0081) & \textbf{0.0140(.0020)} & -- & 0.0191(.0047) & 0.0147(.0021) \\
\addlinespace[1pt]
Segmentation & Final all mIoU $\uparrow$ & 0.0917(.0061) & 0.0946(.0040) & 0.3080(.0307) & \textbf{0.4678(.0568)} & 0.4274(.0574) \\
& Base forgetting $\downarrow$ & 0.7135(.0114) & 0.7131(.0166) & 0.4471(.0447) & \textbf{0.2418(.0965)} & 0.2925(.0839) \\
\addlinespace[1pt]
Diffusion & Final denoise MSE $\downarrow$ & 0.0676(.0029) & 0.0296(.0013) & -- & 0.0298(.0014) & \textbf{0.0293(.0013)} \\
& MSE forgetting $\downarrow$ & 0.0503(.0002) & \textbf{-0.0006(.0020)} & -- & 0.0001(.0024) & -0.0003(.0020) \\
\bottomrule
\end{tabular}
\end{table}

\begin{table}[H]
\caption{Configurations for the multi-domain study. ``DR'' denotes EWC-DR.}
\label{tab:empirical-protocol}
\centering
\small
\begin{tabular}{@{}p{0.13\linewidth}p{0.38\linewidth}p{0.43\linewidth}@{}}
\toprule
Task type & Dataset, sequence, and model & Selected values \\
\midrule
Classification & ImageNet-R; ten 20-class distributions; pretrained ViT-B/16 with LoRA rank 4. & OPG: $\lambda=333.333$; DR: $\lambda=3$; U-IEWC: $\lambda=33.333,\tau=10^{-4}$; weighted IEWC: $\lambda=11.111,\tau=10^{-2}$. \\
NLP & TRACE, C-STANCE $\rightarrow$ FOMC $\rightarrow$ ScienceQA; pretrained Qwen2.5-3B with LoRA rank 8. & OPG: $\lambda=1$; DR: $\lambda=1$; U-IEWC: $\lambda=0.111,\tau=10^{-2}$; weighted IEWC: $\lambda=59049,\tau=10^{-2}$. \\
Forecasting & Weather $\rightarrow$ ECL $\rightarrow$ Traffic; PatchTST trained from scratch. & OPG: $\lambda=729$; U-IEWC: $\lambda=9,\tau=10^{-2}$; weighted IEWC: $\lambda=9,\tau=10^{-2}$. \\
Segmentation & VOC 15--5, base then novel classes; pretrained ConvNeXtV2-tiny FPN with full fine-tuning. & OPG: $\lambda=0.111$; DR: $\lambda=243$; U-IEWC: $\lambda=270,\tau=10^{-3}$; weighted IEWC: $\lambda=2430,\tau=10^{-2}$. \\
Diffusion & MNIST DDPM, labels $0/1\rightarrow2/3$; conditional DDPM trained from scratch. & OPG: $\lambda=9\cdot10^5$; U-IEWC: $\lambda=675,\tau=10^{-3}$; weighted IEWC: $\lambda=225,\tau=10^{-3}$; output ablation: $\lambda=10^6,\tau=10^{-3}$. \\
\bottomrule
\end{tabular}
\end{table}

The multi-domain search used a five-point multiplicative grid around a task-specific midpoint and extended the grid up to three times when an endpoint was best. Its results therefore report the best-performing configuration in each setting. The held-out evaluation instead fixes all five grid points before validation, never extends the grid, and accesses the test split only after the selected hyperparameters are saved.

\begin{figure}[H]
\centering
\input{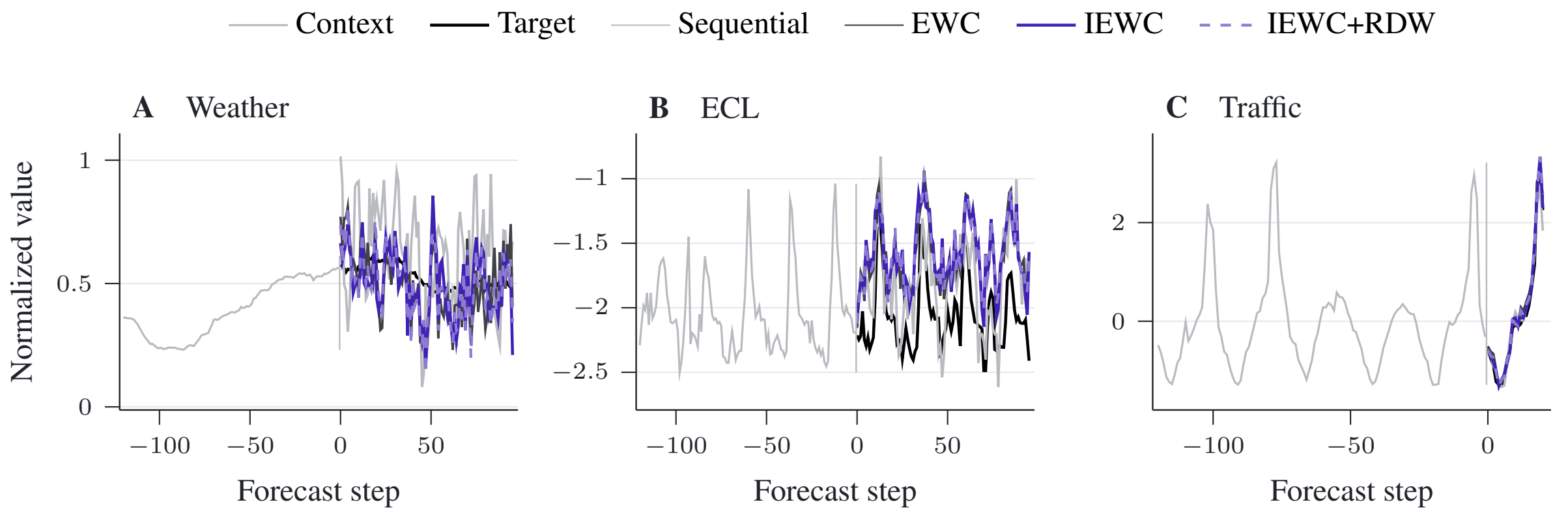}
\caption[Forecasting rollout comparison]{Held-out forecasting rollouts after the continual sequence. Gray: observed context; black: target horizon; colored: method predictions.}
\label{fig:forecast-rollouts}
\end{figure}

The archive also contains all 635 per-$\lambda$ tuning runs behind the five reported settings: 148 vision-classification, 161 forecasting/regression, 47 DDPM, 55 segmentation, 110 encoder-language, and 114 TRACE decoder-only runs. They span CIFAR-100 and ImageNet-R; ETT, M4, and long-horizon forecasting; synthetic and GLUE-family language sequences; three Qwen scales on TRACE; VOC 15--5; and MNIST DDPM, with sequential, OPG-EWC, EWC-DR, uniform IEWC, residual-diversity, and FROMP-derived variants where implemented.

\subsection{Damping sensitivity}

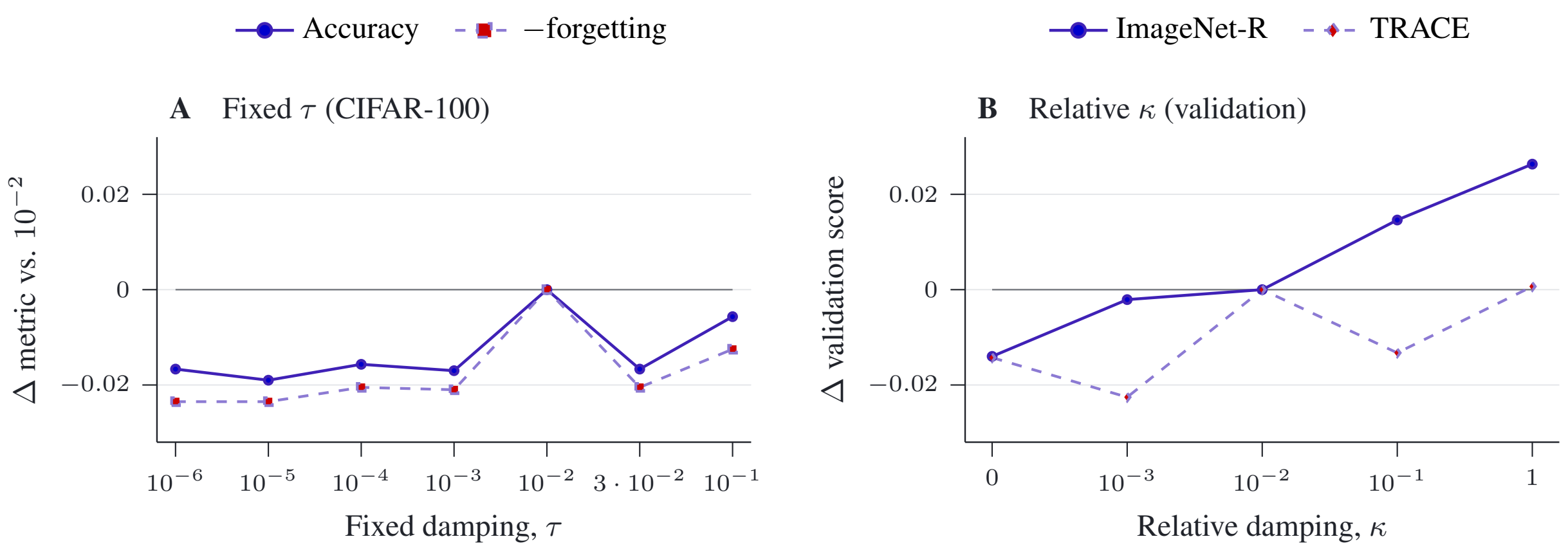
\begin{figure}[H]
\centering
\begin{tikzpicture}
\begin{groupplot}[group style={group size=2 by 1,horizontal sep=22mm},iewc axis,width=.37\linewidth,height=.19\linewidth,ymin=-.032,ymax=.032]
\nextgroupplot[title={\textbf{A}\quad Fixed $\tau$ (CIFAR-100)},xlabel={Fixed damping, $\tau$},ylabel={$\Delta$ metric vs. $10^{-2}$},xmin=-.2,xmax=6.2,xtick={0,1,2,3,4,5,6},xticklabels={$10^{-6}$,$10^{-5}$,$10^{-4}$,$10^{-3}$,$10^{-2}$,$3\cdot10^{-2}$,$10^{-1}$},legend to name=dampingleftlegend,legend columns=2]
\addplot[IEWCMid,line width=.45pt,forget plot] coordinates {(0,0)(6,0)};
\addplot+[accuracy] coordinates {(0,-0.016667) (1,-0.019) (2,-0.015667) (3,-0.017) (4,0) (5,-0.016667) (6,-0.005667)};
\addlegendentry{Accuracy}
\addplot+[retention] coordinates {(0,-0.0235) (1,-0.0235) (2,-0.0205) (3,-0.021) (4,0) (5,-0.0205) (6,-0.0125)};
\addlegendentry{$-$forgetting}
\nextgroupplot[title={\textbf{B}\quad Relative $\kappa$ (validation)},xlabel={Relative damping, $\kappa$},ylabel={$\Delta$ validation score},xmin=-.2,xmax=4.2,xtick={0,1,2,3,4},xticklabels={$0$,$10^{-3}$,$10^{-2}$,$10^{-1}$,$1$},legend to name=dampingrightlegend,legend columns=2]
\addplot[IEWCMid,line width=.45pt,forget plot] coordinates {(0,0)(4,0)};
\addplot+[domainone] coordinates {(0,-0.014001873) (1,-0.0020895897) (2,0) (3,0.014610691) (4,0.026333441)};
\addlegendentry{ImageNet-R}
\addplot+[domaintwo] coordinates {(0,-0.014260163) (1,-0.022556911) (2,0) (3,-0.013260163) (4,0.00066666667)};
\addlegendentry{TRACE}
\end{groupplot}
\node[anchor=south] at ($(group c1r1.north west)!0.5!(group c1r1.north east)+(0,7mm)$) {\pgfplotslegendfromname{dampingleftlegend}};
\node[anchor=south] at ($(group c2r1.north west)!0.5!(group c2r1.north east)+(0,7mm)$) {\pgfplotslegendfromname{dampingrightlegend}};
\end{tikzpicture}
\caption[Damping sensitivity]{Damping sensitivity. \textbf{A}: Fixed-$\tau$ sweep on a CIFAR-100 prefix, expressed relative to $\tau=10^{-2}$; for comparability, forgetting is sign-reversed so higher is better. \textbf{B}: Validation-only task-relative damping at fixed selected $\lambda$, expressed relative to the default $\kappa=10^{-2}$. Panel B uses one validation repetition per domain and was not used for selection.}
\label{fig:damping-sensitivity}
\end{figure}

Figure~\ref{fig:damping-sensitivity} combines two complementary damping diagnostics on a common higher-is-better change scale. Panel A shows a fixed-$\tau$ sweep on a CIFAR-100/FACIL prefix as changes from $\tau=10^{-2}$: the two curves are final accuracy and negative forgetting, so positive values indicate improvement on either measure. Panel B reports a relative-$\kappa$ sweep, fixed in advance, at the selected $\lambda$ on one validation repetition per domain, relative to the default $\kappa=10^{-2}$ of the held-out evaluation. In these single runs, $\kappa=1$ changes validation performance by $+0.0263$ on ImageNet-R and $+0.0007$ on TRACE, so $\kappa$ is worth tuning per domain; the held-out evaluation uses the default throughout. Fixed $\tau$ has loss-dependent units; relative $\kappa$ is dimensionless and preserves global loss-unit invariance.

An additional EWC-DR check tested $\lambda\in\{100,300,1000,3000,10000\}$ on the same CIFAR-100 prefix and selected $100$; it is not part of the ImageNet-R held-out selection.

\section{Label-Noise Stress Test}
\label{app:label-noise}

The CIFAR-100 label-noise experiment provides a downstream stress test alongside the direct matrix analysis in Appendix B. The table reports the best-performing configurations. Uniform training-label flips create atypical high-loss samples, and the clean task-aware test set measures downstream retention. Across this stress test, IEWC loses less accuracy than observed-label OPG-EWC as the flip rate increases.

\begin{table}[H]
\caption{Final average task-aware accuracy under uniform training-label flips, reported as mean $\pm$ across-seed standard deviation.}
\label{tab:label-noise-performance}
\centering
\begin{tabular}{lrrr}
\toprule
Method & $p=0$ & $p=0.1$ & $p=0.25$ \\
\midrule
OPG-EWC & 0.5617 $\pm$ 0.0062 & 0.5132 $\pm$ 0.0112 & 0.2773 $\pm$ 0.0320 \\
EWC-DR & 0.5932 $\pm$ 0.0296 & 0.5310 $\pm$ 0.0226 & 0.4598 $\pm$ 0.0095 \\
IEWC & 0.5978 $\pm$ 0.0260 & 0.5642 $\pm$ 0.0215 & 0.5322 $\pm$ 0.0239 \\
\bottomrule
\end{tabular}
\end{table}

The multi-domain study also compared sorted diagonal entries on a CIFAR-100 prefix. Because that diagnostic does not separate clean and corrupt sample contributions, Appendix~\ref{app:contamination} supplies the direct per-sample full-matrix test; the diagonal summary complements it as a downstream view. It uses all 512 exported entries per matrix and reports the stored-tail mass ratio, normalized stored-tail profile change, and log-scale RMSE of the sorted entries.

\begin{table}[H]
\caption{Sorted-diagonal diagnostics under label noise.}
\label{tab:label-noise-diagonal-diagnostics}
\centering
\begin{tabular}{lrrrrr}
\toprule
Method & Noise & Entries & Tail mass ratio & Profile L1 & Log-scale RMSE \\
\midrule
OPG-EWC & 0.1 & 512 & 0.0288 & 0.2610 & 1.4660 \\
OPG-EWC & 0.25 & 512 & 0.0244 & 0.2737 & 1.5337 \\
IEWC & 0.1 & 512 & 1.0833 & 0.0350 & 0.0350 \\
IEWC & 0.25 & 512 & 1.1003 & 0.0154 & 0.0415 \\
\bottomrule
\end{tabular}
\end{table}

\begin{figure}[H]
\centering
\input{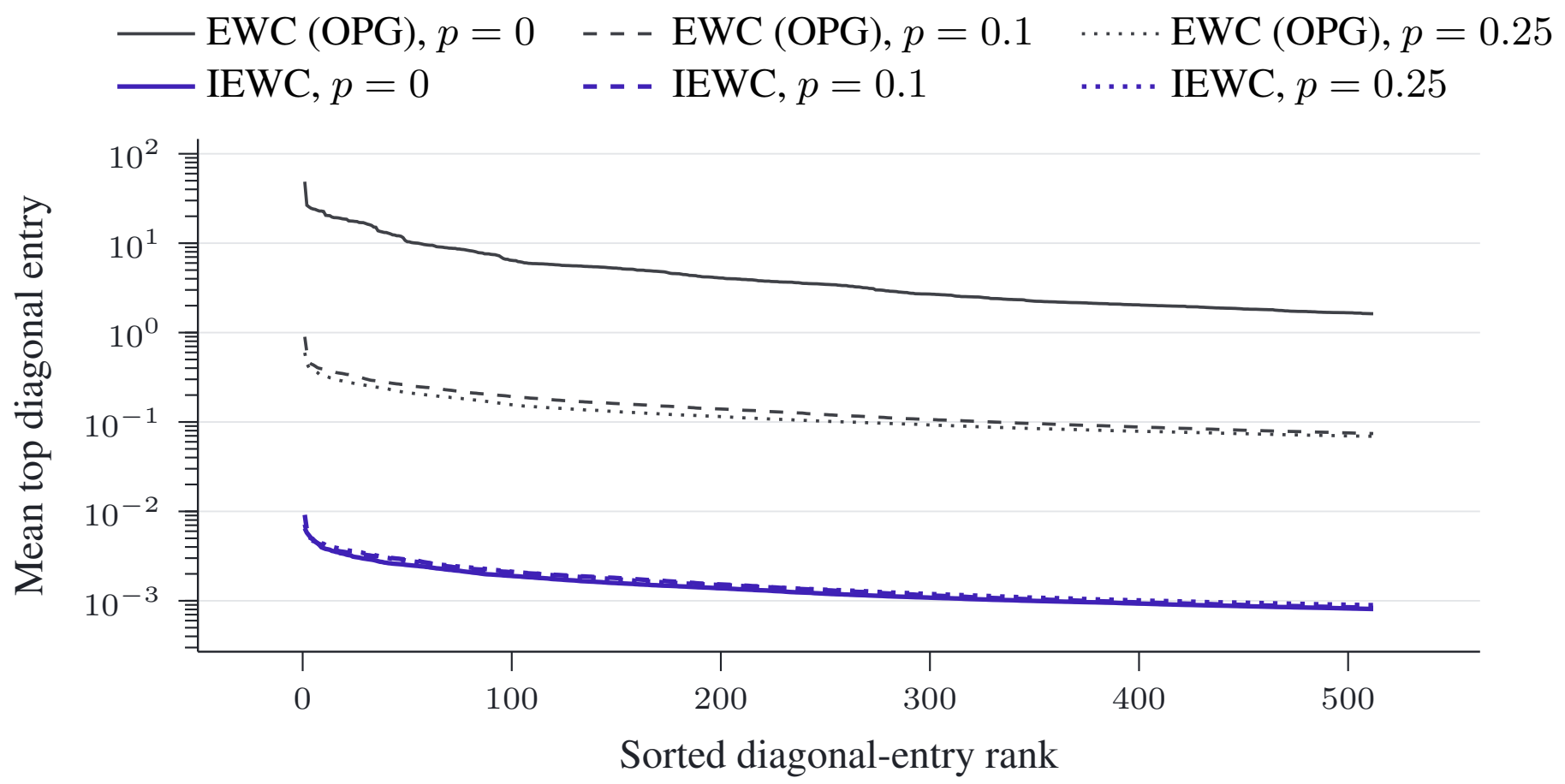}
\caption[Label-noise diagonal diagnostic]{Label-noise diagonal diagnostic. The plot visualizes the sorted high-importance coordinate tail and complements the direct clean/corrupt contribution analysis in Appendix B.}
\label{fig:contamination-diagonal-spectra}
\end{figure}

\section{Reproducibility and Artifact Contents}
\label{app:reproducibility}

The supplementary archive supports two reproduction paths. First, all paper
tables, intervals, and figures can be regenerated on CPU from the saved result
records, without downloading benchmark data or retraining models. Second, the
model results can be rerun from the archived source after obtaining the
datasets and model checkpoints under their original terms. The archive
contains the source snapshot used for all runs with its Git history, a later
snapshot used for reproduction, the manuscript source, the evaluation protocol, the selected hyperparameters,
validation and test records, sensitivity records, mechanism diagnostics,
reporting scripts, a pinned dependency lockfile, and SHA-256 checksums for
every file. A verification script checks file counts, paths, hashes, and source
commits, and confirms that no restricted dataset examples are included.

The empirical scripts use one \texttt{IEWCConfig} interface across the diagonal
and low-rank implementations. The same interface records output geometry,
damping, aggregation weights, and matrix approximation choices, and is used by
both the experiment and reproduction entry points.

\paragraph{Environment and execution.}
All held-out evaluation records come from source commit
\texttt{5e170ea7d84d7094dd1734ee1516b9fba2133a63}. The environment is
reconstructed with \texttt{uv sync --frozen --dev}, and the source passes 75 CPU
tests. Runs enable deterministic PyTorch algorithms, disable cuDNN benchmarking
and TF32, and separately record model, data-loader, importance-order, and
model-Fisher sampling seeds. Each task reseeds training before constructing its
data loader, so method-specific importance computation cannot change a later
task's minibatch order. Hyperparameter selection rejects missing or duplicate
runs and runs from mismatched commits, and test examples are loaded only after
the selected hyperparameters have been saved, with every test access logged.

The held-out evaluation comprises 192 validation runs, 176 test runs, and 48
validation-only sensitivity runs. ImageNet-R uses ten paired class orders.
TRACE uses all six permutations of its three tasks, crossed with two optimizer
seeds; the analysis additionally averages within task order so that the twelve
runs are not treated as twelve independent order draws. The contamination
experiment uses twenty independent seeds at each of four corruption levels.
Every planned run is reported regardless of the direction of its effect. The
analysis reports paired differences, Student-$t$ intervals, exact sign-flip
tests where enumerable, a one-point practical threshold fixed in advance, and
all Holm-corrected secondary comparisons.

\paragraph{Data and model provenance.}
The archive redistributes no benchmark images, language examples, or pretrained
weights. ImageNet-R is obtained from its authors; split definitions are pinned
to Mammoth commit \texttt{e75a491c69fd729edeb01431afb753d9157d9a81}.
The splits contain 19,206 training, 6,000 validation, and 4,794 test images
across 200 classes. The ViT-B/16 checkpoint is pinned to Hugging Face
revision \texttt{063c6c38a5d8510b2e57df480445e94b231dad2c}. TRACE is obtained
from its authors under the Apache-2.0 repository license. Its distributed
train, evaluation, and test files define the three splits for C-STANCE,
ScienceQA, and NumGLUE-cm; their hashes are recorded, and the splits share no
prompts or prompt--answer pairs. FOMC is excluded because its distributed
evaluation and test files are identical. Qwen2.5-3B-Instruct is pinned to
revision \texttt{aa8e72537993ba99e69dfaafa59ed015b17504d1} and remains subject
to its model terms. Mammoth is MIT licensed, and the pinned \texttt{timm}
checkpoint declares Apache-2.0. The IEWC source is supplied for review;
the submission does not override third-party terms or redistribute their
assets.

\paragraph{Compute and complexity.}
The held-out experiments ran on one workstation with an Intel Core
i9-13900K CPU and one NVIDIA GeForce RTX 5090 with 32~GiB of memory. Every run
records total wall time, and every importance phase records wall time and peak
CUDA allocation. With $d$ trainable parameters and $N$ importance samples,
diagonal uniform OPG and IEWC store $O(d)$ values and require $N$
per-sample gradients. RDW maintains at most $K$ unit residual directions on
CPU, using $O(Kd)$ additional storage and $O(NKd)$ inner-product work before
the same diagonal aggregation. The archive includes a table of the measured
cost of every method in both domains.

\paragraph{Regeneration checks.}
The reproduction guide gives exact commands to rebuild the held-out analysis
and all generated tables and figures in a fresh directory. Their checksums are
identical across rebuilds once volatile PDF timestamps are removed. A
clean-checkout test reconstructs the environment, runs the test suite,
regenerates the analysis and paper figures, compiles the manuscript, and
compares the resulting checksums. For the multi-domain study, the archive
includes all 635 tuning runs with per-file checksums;
records that may contain third-party prompts or targets are not
redistributed.

\section{Matched Ordered-CDF Output Geometry}
\label{app:ordered-cdf-geometry}

Consider probability vectors on an ordered grid with $m$ bins. Let
\[
P=I-\frac{1}{m}\mathbf 1\mathbf 1^\top
\]
project onto the probability-simplex tangent, let
$C\in\mathbb R^{(m-1)\times m}$ compute cumulative sums through the
penultimate bin, and let $\Delta x$ be the grid spacing. Define
\[
G_{\mathrm{CDF}}=\Delta x\,P C^\top C P.
\]
For a zero-sum probability perturbation $\Delta z$,
\[
\Delta z^\top G_{\mathrm{CDF}}\Delta z
=
\Delta x\,\|C\Delta z\|_2^2,
\]
which is the squared $L^2$ distance between the corresponding
one-dimensional cumulative distribution functions. Because
$G_{\mathrm{CDF}}$ is singular in the ambient space, its use is restricted to
the probability-simplex tangent. The matched dual form for a loss covector
$r$ is
\[
(Pr)^\top G_{\mathrm{CDF}}^+(Pr),
\]
where $G_{\mathrm{CDF}}^+$ is the Moore--Penrose pseudoinverse. The
implementation tests both the primal identity and the matched dual
construction numerically.

This is an ordered-CDF or Cram\'er metric. It supplies one explicit
space-dependent geometry compatible with IEWC's matched primal/dual
construction. The diffusion experiment uses a separate row/column
cumulative-distribution output representation.

\section{Proof of the Well-Fitted Scaling Proposition}
\label{app:well-fitted-proof}

For $s_n>0$, define
\[
\hat r_n^{,G}
=
\frac{r_n}{\sqrt{s_n}},
\qquad
b_n^{,G}
=
\|J_n^\top\hat r_n^{,G}\|_2^2.
\]
Direct substitution gives
\[
S_{n,0}^{\mathrm{IEF},G}
=
J_n^\top\hat r_n^{,G}(\hat r_n^{,G})^\top J_n,
\qquad
S_n^{\mathrm{OPG}}
=
s_nS_{n,0}^{\mathrm{IEF},G},
\]
and therefore
\[
\|S_{n,0}^{\mathrm{IEF},G}\|_2=b_n^{,G},
\qquad
\|S_n^{\mathrm{OPG}}\|_2=s_nb_n^{,G}.
\]
Since $G_n\preceq M_GI$, we have
\[
\|\hat r_n^{,G}\|_2^2
=
\frac{\|r_n\|_2^2}{r_n^\top G_n^{-1}r_n}
\le M_G.
\]
The bounded-Jacobian assumption consequently gives
$b_n^{,G}\le C_J^2M_G$, which proves the two upper bounds. The
nondegeneracy condition $\|J_n^\top\hat r_n^{,G}\|_2\ge c_J$ gives
the corresponding lower bounds.

For Euclidean softmax cross-entropy with correct-class probability
$p_{n,c}=1-\varepsilon$, the non-target probabilities sum to
$\varepsilon$, and hence
\[
\varepsilon^2
\le
\|p_n-y_n\|_2^2
=
\varepsilon^2+\sum_{k\ne c}p_{n,k}^2
\le
2\varepsilon^2.
\]
Thus $s_n\in\Theta(\varepsilon^2)$, which gives the stated rates under
the same nondegeneracy condition.

For positive damping, multiplying and dividing by $s_n$ yields
\[
S_{n,\tau}^{\mathrm{IEF},G}
=
\frac{s_n}{s_n+\tau}S_{n,0}^{\mathrm{IEF},G}.
\]
With fixed $\tau>0$, bounded $J_n$ gives
$\|J_n^\top r_n\|_2^2/(s_n+\tau)\to0$ as $r_n\to0$. Finally, under
a global positive loss rescaling by $c$, both $r_n$ and
$g_n=J_n^\top r_n$ scale by $c$, while $s_n$ and the task-relative
$\tau_A$ scale by $c^2$. Their quotient is therefore unchanged.

\section{Additional Output-Space Derivations}

\subsection{Matched-metric hyperplane projection}
\label{app:matched-metric-projection}

For the tangent hyperplane $T=\{t:r^\top t=0\}$, write the normal residual as $e=\Delta z-t$ with $t\in T$. The projection solves
\[
\min_e e^\top G e
\quad\text{subject to}\quad
r^\top e=r^\top\Delta z.
\]
The Lagrange equations give
\[
e^\star
=
\frac{r^\top\Delta z}{r^\top G^{-1}r}G^{-1}r,
\]
whose squared $G$-norm is
\[
\|e^\star\|_G^2
=
\frac{(r^\top\Delta z)^2}{r^\top G^{-1}r}.
\]
Hence $G$ measures primal output displacement and $G^{-1}$ acts on the loss covector.

\subsection{Scalar least-squares reduction}
\label{app:scalar-gauss-newton}

Consider scalar-output least-squares regression in Euclidean geometry,
\[
\ell_n(z)=\frac12(z-y_n)^2.
\]
Then $r_n=z_n^A-y_n$ is scalar and, for $r_n\neq0$, the undamped normalized factor is $r_n^2/r_n^2=1$. Each IEF summand therefore reduces to $J_n^\top J_n$, and averaging gives
\[
M_{\mathrm{IEF}}
=
\frac{1}{N}\sum_{n=1}^N J_n^\top J_n.
\]
This is exactly the Gauss--Newton or output-Jacobian form for squared-error regression \citep{GaussNewton,MartensHF,MartensNG}, as also made explicit by \citet{IEF}.

\section{Notation}
\label{app:notation}

\begin{table}[H]
\caption{Notation and distinctions used throughout the paper.}
\label{tab:notation}
\centering
\small
\begin{tabular}{@{}p{0.18\linewidth}p{0.76\linewidth}@{}}
\toprule
Symbol & Meaning \\
\midrule
$g_i=J_i^\top r_i$ & Parameter gradient of the selected scalar functional; IEWC uses the old-sample loss. \\
$M_{\mathrm{OPG}}$ & Mean $g_i g_i^\top$ from scalar-loss preservation; only the likelihood-score case is the observed-label empirical-Fisher algebra. \\
$F_{\mathrm{model}}$ & Fisher information averaging score outer products over labels sampled from the frozen predictive distribution. \\
$s_i=r_i^\top G_i^{-1}r_i$ & Squared dual norm of the output-loss gradient under the matched output metric. \\
$M_{\mathrm{IEF}}$ & Level-set-normalized mean $g_i g_i^\top/(s_i+\tau_A)$; the undamped Euclidean case is iEF. \\
$\rho_\tau(s)$ & Damping attenuation $s/(s+\tau)$ relative to the undamped summand. \\
$\kappa$ & Dimensionless relative-damping coefficient in $\tau_A=\kappa\operatorname{median}_{s_i>0}s_i$. \\
$\bar h_i$ & Optional mean-one aggregation weight; uniform weighting and RDW are evaluated separately. \\
$(\Gamma,h,\Pi)$ & Output-metric rule, aggregation-weight rule, and matrix approximation operator. \\
\bottomrule
\end{tabular}
\end{table}

\end{document}